\documentclass[letterpaper]{article} 
\usepackage{aaai24}  
\usepackage{times}  
\usepackage{helvet}  
\usepackage{courier}  
\usepackage[hyphens]{url}  
\usepackage{graphicx} 
\def\UrlFont{\rm}  
\usepackage{natbib}  
\usepackage{caption} 
\usepackage{algorithm}
\usepackage{algorithmic}

\usepackage{subfig} 
\usepackage{caption}
\usepackage{booktabs}
\usepackage{amsmath}
\usepackage{amsfonts}
\usepackage{multirow}

\usepackage{newfloat}
\usepackage{listings}
\DeclareCaptionStyle{ruled}{labelfont=normalfont,labelsep=colon,strut=off} 
\floatstyle{ruled}
\newfloat{listing}{tb}{lst}{}
\floatname{listing}{Listing}
\usepackage[table]{xcolor}

\makeatletter
\renewcommand{\maketag@@@}[1]{\hbox{\m@th\normalsize\normalfont#1}}%
\makeatother

\title{Three Heads Are Better Than One: \\Complementary Experts for Long-Tailed Semi-supervised Learning}
\author{
    Chengcheng Ma\textsuperscript{\rm 1, 2}\thanks{This work was done during an internship in Huawei Inc.\vspace{-0.5mm}}, Ismail Elezi\textsuperscript{\rm 3}, Jiankang Deng\textsuperscript{\rm 3}, Weiming Dong\textsuperscript{\rm 1}\thanks{Corresponding author (weiming.dong@ia.ac.cn).}, Changsheng Xu\textsuperscript{\rm 1}
}
\affiliations{
    \textsuperscript{\rm 1}Institute of Automation, Chinese Academy of Sciences, Beijing, China\\
    \textsuperscript{\rm 2}School of Artificial Intelligence, University of Chinese Academy of Sciences, Beijing, China\\
    \textsuperscript{\rm 3}Huawei Noah’s Ark Lab, London, UK

}

\usepackage{bibentry}

\begin{document}

\maketitle

\begin{abstract}
We address the challenging problem of Long-Tailed Semi-Supervised Learning (LTSSL) where labeled data exhibit imbalanced class distribution and unlabeled data follow an unknown distribution.
Unlike in balanced SSL, the generated pseudo-labels are skewed towards \textit{head} classes, intensifying the training bias.
Such a phenomenon is even amplified as more unlabeled data will be mislabeled as \textit{head} classes when the class distribution of labeled and unlabeled datasets are mismatched.
To solve this problem, we propose a novel method named ComPlementary Experts (CPE). 
Specifically, we train multiple experts to model various class distributions, each of them yielding high-quality pseudo-labels within one form of class distribution.
Besides, we introduce Classwise Batch Normalization for CPE 
to avoid performance degradation caused by feature distribution mismatch between head and non-head classes.
CPE achieves state-of-the-art performances on CIFAR-10-LT, CIFAR-100-LT, and STL-10-LT dataset benchmarks. For instance, on CIFAR-10-LT, CPE improves test accuracy by over 2.22\% compared to baselines.
Code is available at \url{https://github.com/machengcheng2016/CPE-LTSSL}.
\end{abstract}
\section{Introduction}
\label{sec: introduction}

Semi-supervised learning (SSL) is a promising approach for improving the performance of deep neural networks (DNNs). 
In SSL, only a limited amount of labeled data is available, with the majority of data being unlabeled. 
Most existing SSL approaches assume that the labeled and unlabeled data follow the same distribution and that the distribution is class-balanced. 
However, this assumption is often violated in real-world applications, where the class distribution of the labeled data is often long-tailed, with a few (head) classes containing significantly more samples than the other (tail) classes.
In long-tailed semi-supervised learning (LTSSL), the distribution of the unlabeled data can take three forms: (i) it can be \textit{long-tailed}, following the same distribution as the labeled data; (ii) it can be \textit{uniform}, with a balanced class distribution; or (iii) it can be \textit{inverse long-tailed}, with the head classes of the labeled data being the tail classes of the unlabeled data, and vice versa.

In the fully-supervised setting, DNNs trained on imbalanced labeled data are prone to mostly predict the head classes. 
This issue is exacerbated in LTSSL, as the unlabeled data are mostly pseudo-labeled as head classes during training, intensifying the training bias. 
This can lead to lower test accuracy on tail classes, especially when the class distributions of the labeled and unlabeled data are mismatched.
Unfortunately, most of the pioneer LTSSL works assume a similar class distribution between labeled and unlabeled sets, causing lower performances when such assumption is inevitably violated. 
Very recently, ACR~\cite{ACR} proposes to handle varying class distributions of unlabeled sets by introducing adaptive logit adjustment and achieving state-of-the-art (SOTA) performance.
Though effective, its pseudo-labels are generated by a single classifier. 
As we will show in the experimental section, the pseudo-label quality by ACR can be further improved.

\begin{figure*}[t]
    \centering
    \subfloat[\textit{Consistent} case]{\label{fig:F1-E1-crop}
    {\includegraphics[width=0.31\textwidth]{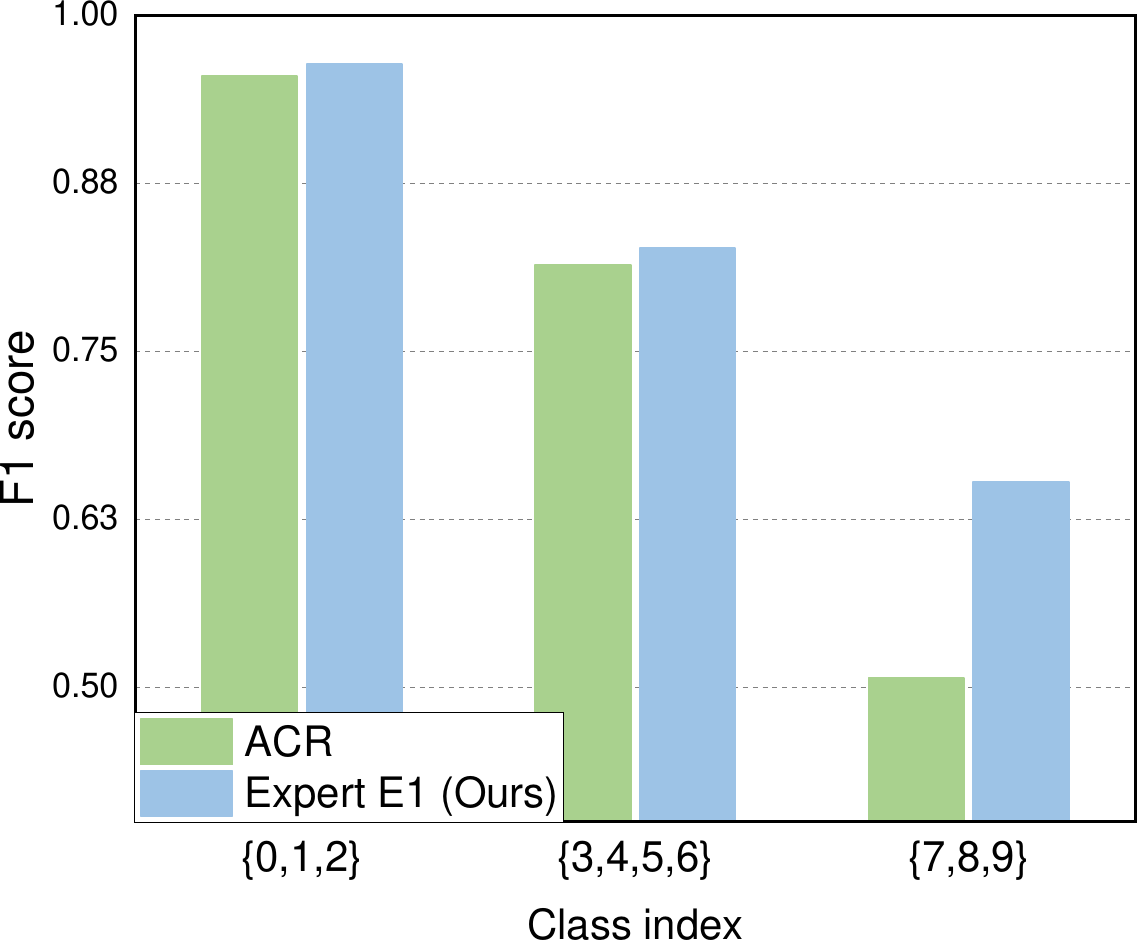}}}\hfill
    \subfloat[\textit{Uniform} case]{\label{fig:F1-E2-crop}
    {\includegraphics[width=0.31\textwidth]{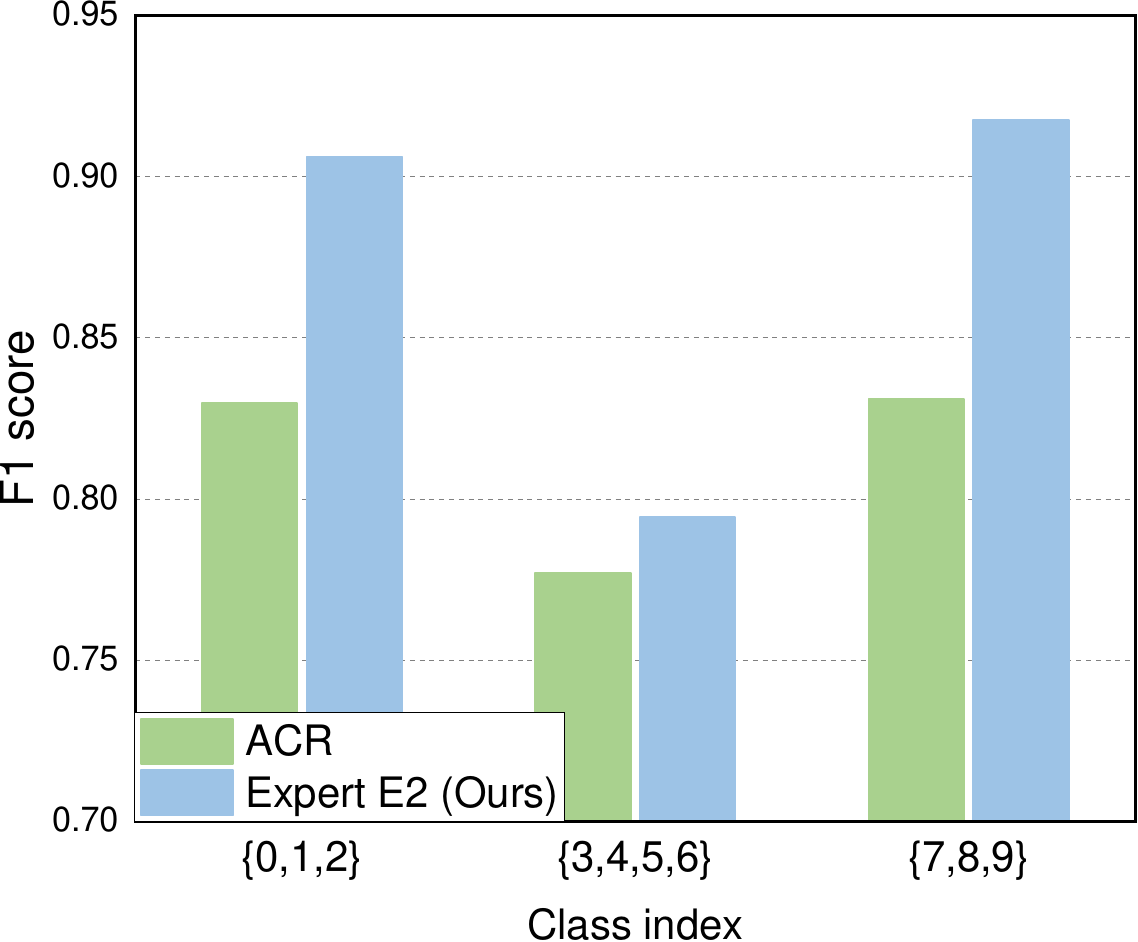}}}\hfill
    \subfloat[\textit{Inverse} case]{\label{F1-E3-crop}
    {\includegraphics[width=0.31\textwidth]{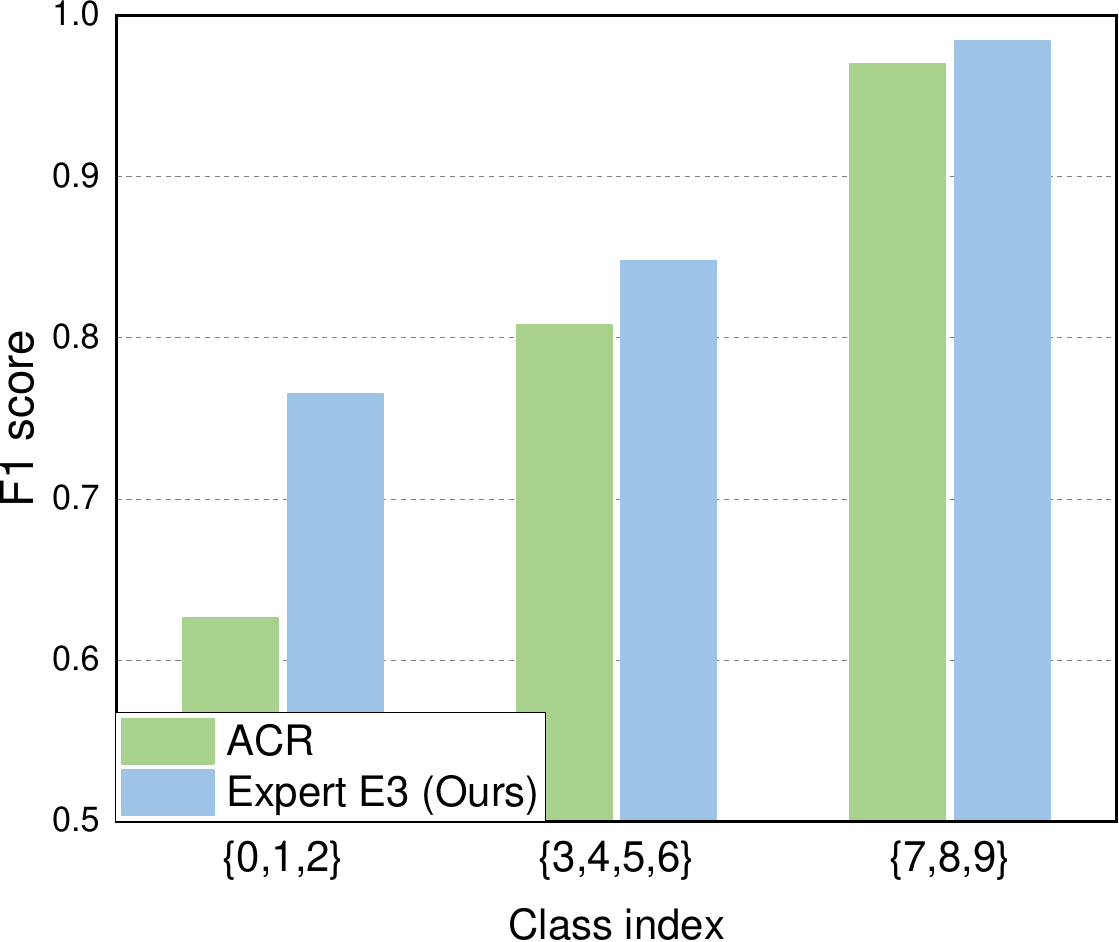}}}\hfill\hfill
    \caption{
    Comparison of F1 score of pseudo-label predictions between ACR~\cite{ACR} and our CPE method under the \{``\textit{consistent}", ``\textit{uniform}", ``\textit{inverse}"\} cases. 
    The dataset is CIFAR-10-LT with imbalance ratio $\gamma_l$ being 100. 
    One of experts in our CPE can generate pseudo-labels with higher quality.
    }
    \label{fig:pseudo-label-F1}
\end{figure*}

In this paper, we propose a flexible and end-to-end LTSSL algorithm, namely \textbf{C}om\textbf{P}lementary \textbf{E}xperts (CPE).
To cope with varying class distributions of unlabeled sets, we train multiple experts where each expert is tasked to model one class group.
We do so by using different logit adjustments~\cite{LA} during the computation of the supervised loss.
Specifically, the first expert is trained with regular cross entropy loss to fit the skewed class distribution of labeled sets, so it is natural to see it perform well when the distribution of unlabeled and labeled sets are consistent.
In contrast, the second and third experts are trained with different intensities of logit adjustments, which are encouraged to yield class-balanced and inversely skewed predictions, respectively.
Thus, they can deal with the uniform and inverse long-tailed cases, where the corresponding shape of predictions is just in need.
In such a complementary manner, one of three experts in our CPE algorithms can always predict pseudo-labels with higher F1 scores than ACR (see Fig.~\ref{fig:pseudo-label-F1}).

Having three experts results in a higher recall in the unlabeled samples belonging to medium and tail classes, especially in the \textit{uniform} and \textit{inverse} cases.
Consequently, the extracted features within medium and tail classes tend to follow distribution with larger variances compared with head classes. 
To prevent such phenomenon from harming model performance, 
we propose Classwise Batch Normalization (CBN) mechanism to realize classwise feature processing.
In CBN, we design three different batch normalization layers for three groups of classes, to affect both the forward and backward calculation during the computation of the unsupervised loss.
We will validate the effectiveness of CBN mechanism in the experimental section.

In summary, our \textbf{contribution} is as follows:
\begin{itemize}
    \item We determine that a single classifier cannot model all possible distributions and propose to model the distributions by three experts trained with different logit adjustments, namely ComPlementary Experts (CPE).
    \item To deal with different feature distributions, we propose Classwise Batch Normalization (CBN) mechanism for CPE training. 
    \item We reach the new state-of-the-art performances on three popular LTSSL benchmarks.
\end{itemize}

\section{Related Work}

\subsection{Semi-supervised learning}
The underlying objective of Semi-Supervised Learning (SSL) is to enhance the performance of Deep Neural Networks (DNNs) through the incorporation of supplementary unlabeled training data, particularly in scenarios where the availability of labeled samples is notably constrained.
Until recently, FixMatch~\cite{fixmatch} and its derivative methodologies such as FlexMatch~\cite{flexmatch}, FreeMatch~\cite{freematch}, and SoftMatch~\cite{softmatch} have emerged as among the most efficacious algorithms for the task of classification. 
These methodologies adhere to the foundational principle of self-training~\cite{pise2008survey}, wherein model predictions on unlabeled data are utilized as pseudo-labels during the training process. 
These approaches concentrate on refining the quality of pseudo-labels through strategies such as curriculum learning~\cite{flexmatch}, adaptive thresholding~\cite{freematch}, and sample weighting~\cite{softmatch}.
However, many existing SSL approaches make the presumption of balanced class distributions within both the labeled and unlabeled training sets. 
This assumption, however, tends to be incompatible with real-world scenarios, where the class distribution of natural data frequently adheres to Zipf's law, manifesting a long-tailed distribution~\cite{johnson2019survey,liu2019large,reed2001pareto}. 
As corroborated by subsequent experiments, the performance of conventional FixMatch deteriorates notably, particularly when the unlabeled training set conforms to an inverse class distribution in relation to the labeled set.

\subsection{Long-tailed learning}
Since the class distribution of the training set is usually imbalanced, existing algorithms in long-tailed learning attempt to improve the generalization ability of DNN towards the \textit{medium} and \textit{tail} classes.
Generally, the framework of existing methods can be summarized into four categories:
\textit{Re-sampling}~\cite{chawla2002smote,buda2018systematic} conducts under-sampling on head classes or over-sampling on tail classes to balance the training data; 
\textit{Re-weighting}~\cite{khan2017cost,cui2019class} adjusts the class weight or sample weight to maintain a balanced training process; 
\textit{Ensemble learning}~\cite{RIDE,ACE,SADE,BalPoE} is based on multiple experts (classifier heads) to enhance the representation learning;
\textit{Loss modification}~\cite{cao2019learning,hong2021disentangling} modifies the logit value by margins in either the training or inference stage and the typical work ``logit adjustment"~\cite{LA} which has been proved as the Bayes-optimal solution of the long-tailed problem.
Nonetheless, the current approaches are predominantly tailored for the fully-supervised context, rendering them unsuitable for semi-supervised learning 
where the class distribution is unspecified.

\subsection{Long-tailed Semi-supervised learning}
Due to its practicality, LTSSL has reached increasing attention in recent years.
Different from balanced SSL, the unlabeled data are prone to be pseudo-labeled as head classes during training, leading to low recall on tail classes. 
To address the challenge of LTSSL, previous works draw inspiration from existing solutions to long-tailed classification to generate unbiased and accurate pseudo-labels, such as re-sampling~\cite{ABC,Adsh}, re-weighting~\cite{SAW}, transfer learning~\cite{CoSSL}, and logit adjustment~\cite{DePL}.
However, these works usually assume a similar class distribution between labeled and unlabeled set and show inferior performances when such assumption is violated.
Another branch of LTSSL works follows the popular idea of SSL to augment the labeled training set with the most reliable unlabeled data multiple times~\cite{Crest,Simis}.
However, such approaches are not in an end-to-end fashion and require multiple re-training costs.
Very recently, ACR~\cite{ACR} is proposed to dynamically adjust the intensity of logit adjustment by measuring the distribution divergence between labeled and unlabeled sets, which can handle varying class distributions of unlabeled set and achieves the state-of-the-art (SOTA) performances.
Though effective, its pseudo-labels are generated by one single classifier leading to sub-optimal performance. We will show in the following sections that the pseudo-label quality can still be improved.
\section{Methodology}

\begin{figure*}[t]
\centering
{
\includegraphics[width=0.75\linewidth]{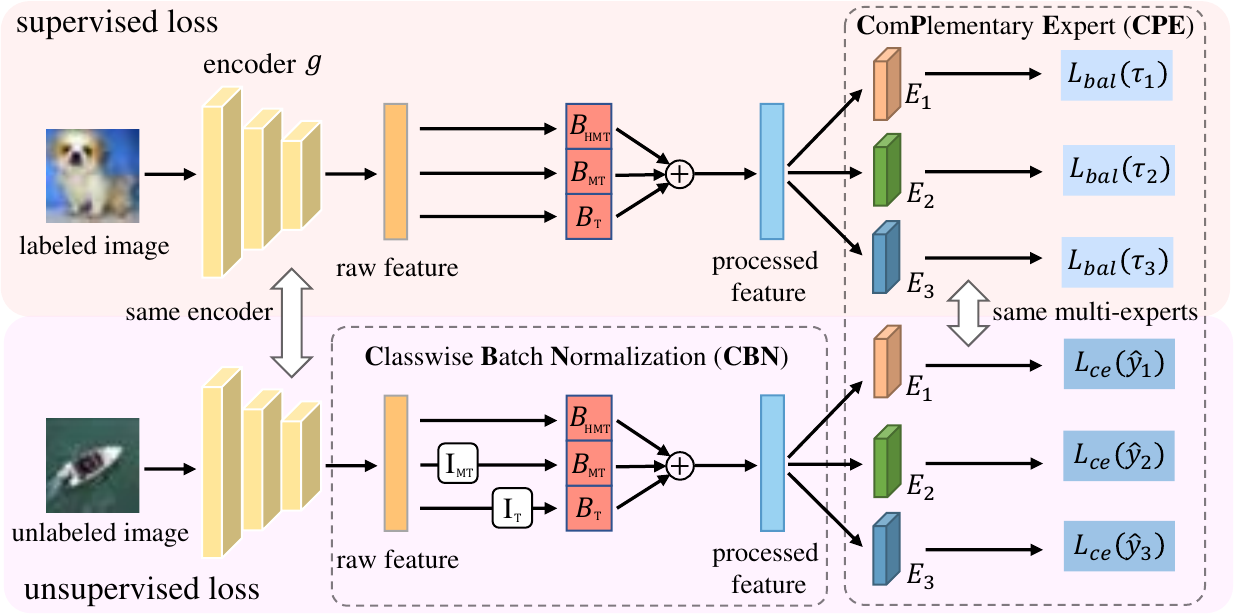}
}
\caption{Overview of our CPE algorithm together with the CBN mechanism.
I$_{\text{MT}}$ and I$_{\text{T}}$ represent the class mask in Eq. (\ref{eq: unsup loss with CBN}). $L_{bal}(\tau)$ is short for balanced cross entropy loss in Eq. (\ref{eq: logit adjustment}). $L_{ce}(\hat{y})$ denotes the cross entropy loss with pseudo-label $\hat{y}$.
} 
\label{fig: overview}
\end{figure*}

\subsection{Preliminaries}
\textbf{Problem Formulation.}\;
Long-tailed semi-supervised learning, also known as imbalanced semi-supervised learning, involves a labeled training set $\mathcal{D}_l=\{\mathbf{x}_i^l,y_i\}_{i=1}^N$ and an unlabeled training set $\mathcal{D}_u=\{\mathbf{x}_j^u\}_{j=1}^M$ sharing the same class set, where $\mathbf{x}$ represents a training sample and $y\in\{1,\dots,C\}$ represents a class label.
Let $N_k$ and $M_k$ be the number of samples within the $k$ class in $\mathcal{D}_l$ and $\mathcal{D}_u$, respectively, so $\sum_{k=1}^{C}N_k=N$ and $\sum_{k=1}^{C}M_k=M$.
As the class distribution of $\mathcal{D}_l$ is skewed, the classes are usually sorted in descending order, \textit{i.e.}, $N_1>N_2>\cdots>N_C$. Correspondingly, the imbalance ratio of $\mathcal{D}_l$ is defined as $\gamma_l=N_1/N_C$.
As for $\mathcal{D}_u$, the class distribution is unknown because all labels are unavailable. In this paper, we follow previous works~\cite{DASO,ACR} to consider three cases of class distribution of unlabeled set, that are \{``\textit{consistent}", ``\textit{uniform}", ``\textit{inverse}"\}.
In \textit{consistent} case, we have $M_1>M_2>\cdots>M_C$ and $\gamma_u=\gamma_l$, where $\gamma_u$ is defined as $M_1/M_C$.
In \textit{uniform} case, we have $M_1=M_2=\cdots=M_C$ and $\gamma_u=1$.
In \textit{inverse} case, we have $M_1<M_2<\cdots<M_C$ and $\gamma_u=1/\gamma_l$.

The goal of LTSSL is to train a DNN classifier $f(\cdot)$ parameterized by $\theta$ on $\mathcal{D}_l$ and $\mathcal{D}_u$, and to achieve good performances on a class-balanced test set $\mathcal{D}_{test}$.
Generally, the training loss of LTSSL is similar to that of SSL and is composed of a supervised term and an unsupervised term:
\vspace{-2mm}
\begin{small}
\begin{align}
L
&=\frac{1}{B_l}\sum_{i=1}^{B_l}\ell_{CE}\big( f(\mathbf{x}_i^l), y_i \big) \nonumber\\
&+ \frac{\lambda}{B_u}\sum_{j=1}^{B_u}\ell_{CE}\big( f(\mathbf{x}_j^u), \hat{y}_j \big) \cdot \mathbb{I}\bigg(\|\sigma\big(f(\mathbf{x}_j^u)\big)\|_\infty > \rho\bigg),
\label{eq: SSL}
\end{align}
\end{small}where $\hat{y}_j$ denotes the pseudo-label of unlabeled sample $\mathbf{x}_j^u$, usually being the model prediction $\hat{y}_j = \arg\max f(\mathbf{x}_j^u)$.
$\sigma(\cdot)$ denotes the softmax function, and $\mathbb{I}(\cdot)$ denotes a binary sample mask to select samples with confidence larger than the threshold $\rho$.
$\ell_{CE}(\cdot, \cdot)$ denotes the cross entropy loss.
$B_l$ and $B_u$ denote the size of labeled and unlabeled data batches, respectively. 
$\lambda$ is a hyper-parameter that controls the strength of the unsupervised loss.

\begin{table}[!t]
\centering
\scriptsize
\renewcommand\arraystretch{1.15}
\scalebox{1.18}{
\begin{tabular}{ccccc}
\hline
\multirow{2}{*}{Distribution case} & \multirow{2}{*}{$\tau$} & \multicolumn{3}{c}{F1 score of pseudo-labels} \\
                                   &                         & Head          & Medium          & Tail       \\\hline
\multirow{3}{*}{\shortstack[c]{\textit{Consistent}\\($\gamma_l=\gamma_u=100$)}}        
                                   & \textbf{0}                       & \textbf{0.96}        & \textbf{0.81}          & \textbf{0.65}     \\
                                   & 2                       & 0.95        & 0.80          & 0.51     \\
                                   & 4                       & 0.28        & 0.80          & 0.27     \\\hline
\multirow{3}{*}{\shortstack[c]{\textit{Uniform}\\($\gamma_l=100,\gamma_u=1$)}}        
                                   & 0                       & 0.88        & 0.79          & 0.87     \\
                                   & \textbf{2}                       & \textbf{0.91}        & \textbf{0.80}          & \textbf{0.92}     \\
                                   & 4                       & 0.28        & 0.79          & 0.73     \\\hline
\multirow{3}{*}{\shortstack[c]{\textit{Inverse}\\($\gamma_l=100,\gamma_u=\frac{1}{100}$)}}        
                                   & 0                       & 0.41        & 0.82          & 0.92     \\
                                   & 2                       & 0.59        & 0.83          & 0.97     \\
                                   & \textbf{4}                       & \textbf{0.77}        & \textbf{0.85}          & \textbf{0.98}     \\\hline
\end{tabular}
}
\vspace{-1mm}
\caption{F1 score of pseudo-labels under three different cases with varying $\tau$ in Eq. (\ref{eq: logit adjustment}). The dataset is CIFAR-10-LT. $\tau$ being 2 means balanced training on labeled set, while $\tau$ being 0 or 4 means under- or over- balancing. We clearly see that each $\tau$ can only generate high-quality pseudo-labels in one distribution case.}
\vspace{-2mm}
\label{tab: each-expert-does-its-work}
\end{table}

\vspace{2mm}
\noindent\textbf{Logit Adjustment for Long-Tailed Recognition.}\;\,
The logit adjustment~\cite{LA} is theoretically proven as the Bayes-optimal solution to the long-tailed recognition task.
Formally, the output logits are added to the logarithm of prior $\pi$ during training, formulating the balanced cross entropy loss:
\vspace{-2mm}
\begin{small}
\begin{align}
L_{bal}= 
\frac{1}{B}\sum_{i=1}^{B}\ell_{CE}\big( f(\mathbf{x}_i) + \tau \cdot \log \pi, \; y_i \big).
\label{eq: logit adjustment}
\end{align}
\end{small}The prior $\pi$ is pre-computed as the label frequencies, \textit{i.e.}, $\pi_i=N_i/N$.
The positive scaling factor $\tau$ controls the intensity of adjustment.
Intuitively, the value of $\log\pi_i$ lies in $(-\infty,0)$, and is close to $-\infty$ for the tail classes.
By adjustment, the logit values of tail classes are encouraged to cancel out the value of $\log\pi_i$ during training, thus becomes larger than those of head classes ultimately.
On the other hand, the DNN classifier 
trained on imbalanced dataset
tends to yield larger logit values for head classes, thus the logit values for all classes can be balanced if classifier is trained by $L_{bal}$, leading to class-balanced predictions during test time.

However, under the LTSSL setting, the unknown class distribution of unlabeled set makes the logit adjustment algorithm somewhat problematic.
To be specific, if we still pre-compute $\pi$ of $\mathcal{D}_l$ and replace $\ell_{CE}$ with $L_{bal}$ in the first term of Eq. (\ref{eq: SSL}),  
then only in the \textit{uniform} case can the classifier generate high-quality pseudo-labels (see Table \ref{tab: each-expert-does-its-work} for an empirical evidence).
This is because $L_{bal}$ always prefers class-balanced predictions.
In contrast, if the class distribution of $\mathcal{D}_u$ is also skewed such as in the \textit{consistent} or \textit{inverse} case, then the classifier needs to be under-balanced or over-balanced to cope with the corresponding case (with lower or higher $\tau$), which contradicts the supervised loss on labeled set in turn.
%
We will show how to address such a dilemma in the next section.

\subsection{ComPlementary Experts (CPE)}
The two-stage training~\cite{decoupling} is a recognized approach for long-tailed fully-supervised learning. 
It involves separate training of the feature extractor and classification head in two stages. 
However, due to its computational demands, in LTSSL, prior works~\cite{CoSSL,ACR} often opt for a single-stage training using a dual-head DNN architecture.
Typically, these algorithms update the shared feature extractor by aggregating gradients from the first head to facilitate representation learning. The second head is then updated using techniques tailored for long-tailed scenarios, aiming to achieve balanced predictions. An example is substituting $\ell_{CE}$ with $L_{bal}$ in the first term of Eq. (\ref{eq: SSL}), which is finally employed for evaluation.

However, we notice that all existing methods only utilize one single head for pseudo-label generation, which cannot perfectly fit the varying class distributions of unlabeled set.
According to Table \ref{tab: each-expert-does-its-work}, the quality of pseudo-labels can be less satisfying if there exists a mismatch between logit adjustment intensity $\tau$ and imbalance ratio $\gamma_u$.
To this end, we propose to concatenate the shared feature extractor $g(\cdot)$ with three different heads, namely experts \{$E_1$, $E_2$, $E_3$\}.
Specifically, each of experts is trained by balanced cross entropy loss $L_{bal}$ (see Eq. (\ref{eq: logit adjustment})) with a specified intensity $\tau_i$, and the total supervised loss is formulated as
\vspace{-2mm}
\begin{small}
\begin{align}
L_{CPE}^{sup}
&=\frac{1}{B_l}\sum_{i=1}^{B_l}\ell_{CE}\bigg(E_1\big(g(\mathbf{x}_i^l)\big)+\tau_1\cdot\log\pi, \;y_i \bigg) \nonumber\\
&+\frac{1}{B_l}\sum_{i=1}^{B_l}\ell_{CE}\bigg(E_2\big(g(\mathbf{x}_i^l)\big)+\tau_2\cdot\log\pi, \;y_i \bigg) \nonumber\\
&+\frac{1}{B_l}\sum_{i=1}^{B_l}\ell_{CE}\bigg(E_3\big(g(\mathbf{x}_i^l)\big)+\tau_3\cdot\log\pi, \;y_i \bigg).
\label{eq: CPE-sup}
\end{align}
\end{small}Correspondingly, the formulation of total unsupervised loss follows that of Eq. (\ref{eq: SSL}):

\begin{small}
\begin{align}
&L_{CPE}^{unsup}\nonumber\\
=&\frac{\lambda}{B_u}\sum_{j=1}^{B_u}\ell_{CE}\bigg( E_1\big(g(\mathbf{x}_j^u)\big), \hat{y}_{j(1)} \bigg) \cdot \mathbb{I}\bigg(\|\sigma\big(E_1\big(g(\mathbf{x}_j^u)\big)\big)\|_\infty > \rho\bigg)\nonumber\\
+&\frac{\lambda}{B_u}\sum_{j=1}^{B_u}\ell_{CE}\bigg( E_2\big(g(\mathbf{x}_j^u)\big), \hat{y}_{j(2)} \bigg) \cdot \mathbb{I}\bigg(\|\sigma\big(E_2\big(g(\mathbf{x}_j^u)\big)\big)\|_\infty > \rho\bigg)\nonumber\\
+&\frac{\lambda}{B_u}\sum_{j=1}^{B_u}\ell_{CE}\bigg( E_3\big(g(\mathbf{x}_j^u)\big), \hat{y}_{j(3)} \bigg) \cdot \mathbb{I}\bigg(\|\sigma\big(E_3\big(g(\mathbf{x}_j^u)\big)\big)\|_\infty > \rho\bigg),
\label{eq: CPE-unsup}
\end{align}
\end{small}where $\hat{y}_{(i)}$ denotes the pseudo-label predicted by $E_i$. 
Then the total loss of CPE is $L_{CPE}=L_{CPE}^{sup}+L_{CPE}^{unsup}$.

In training stage, one of the three experts in CPE can always generate high-quality pseudo-labels for each class, complementing the defects of other two experts and ensuring the quality of representation learning.
In inference stage, CPE assumes the test set is class-balanced and utilizes the second expert $E_2$ for evaluation.
The overview of CPE is illustrated in Fig. \ref{fig: overview}.

\subsection{Classwise Batch Normalization for CPE}
\label{sec: CBN}
\begin{figure}[!t]
\centering
{
\includegraphics[width=.85\linewidth]{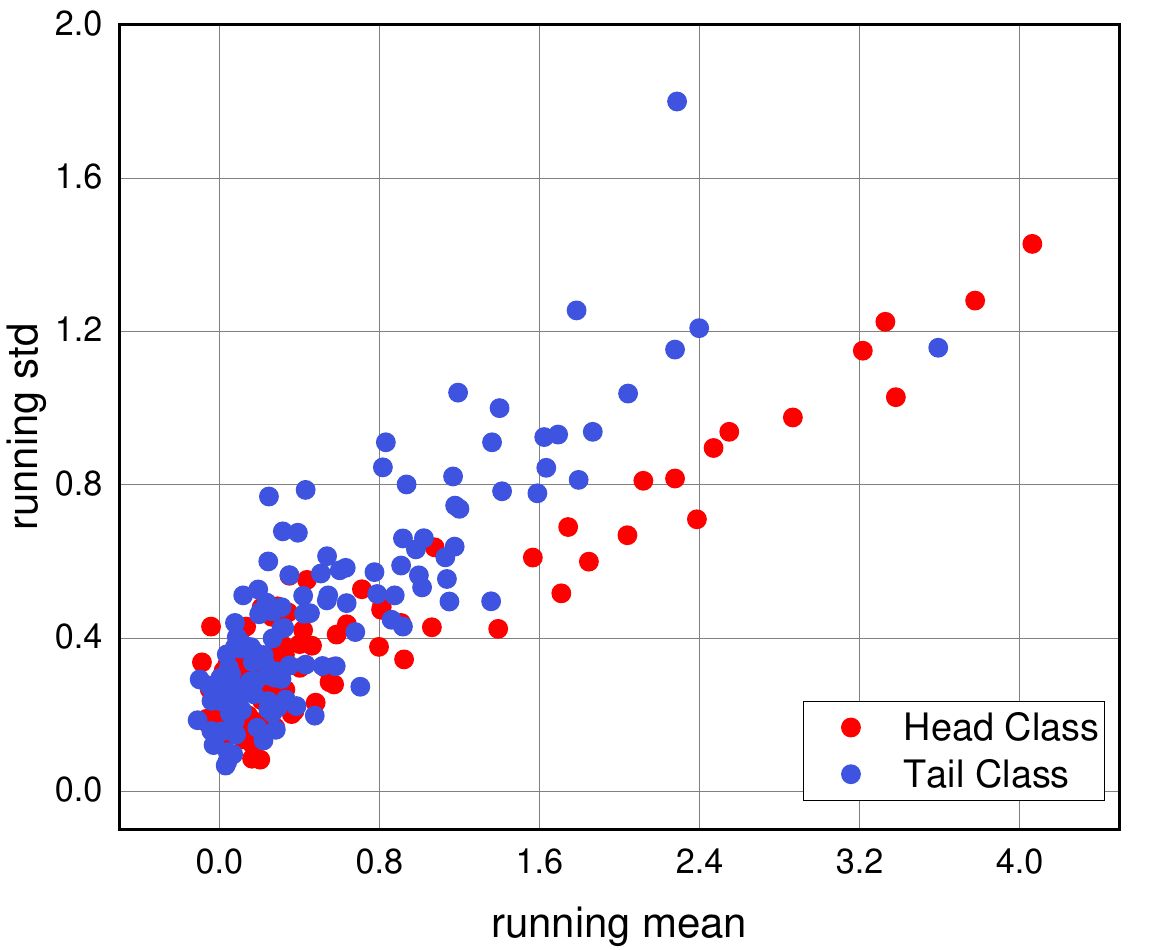}\hspace{2mm}
}
\caption{Statistics of extractor features within head and tail classes in CIFAR-10-LT with $(\gamma_l,\gamma_u)=(100,1/100)$. Each dot represents the running mean and standard deviation of a channel in the BN layer. After pseudo-labeling, features of unlabeled data within tail classes have a higher variance and follow a different distribution from that of the head classes.} 
\vspace{-3mm}
\label{fig: tail-classes-with-higher-variance}
\end{figure}

Despite our CPE method being effective, having three experts results with a higher recall in the unlabeled samples belonging to medium and tail classes, especially in the \textit{uniform} and \textit{inverse} cases.
Consequently, the variance of extracted features in medium and tail classes tends to become larger than that of head classes, as we empirically validate in Fig. \ref{fig: tail-classes-with-higher-variance}.
Specifically, we record the average and standard deviation of each feature channel in the last layer of the feature extractor by class.
%
Such a phenomenon can somehow harm the model performance as previously discussed in adversarial learning~\cite{auxbn} and fully-supervised long-tailed learning~\cite{ResLT}.

To address this issue, inspired by \cite{ResLT}, we propose Classwise Batch Normalization  (CBN) mechanism to realize classwise feature processing.
%
CBN involves three different BN layers to re-scale the features output by the shared encoder $g(\cdot)$, denoted by \{BN$_{\text{HMT}}$, BN$_{\text{MT}}$, BN$_{\text{T}}$\}.
Concretely, BN$_{\text{HMT}}$ processes the features belonging to all classes, while BN$_{\text{MT}}$ processes features in medium and tail classes (denoted by $\mathcal{M}$ and $\mathcal{T}$), and BN$_{\text{T}}$ only processes features in tail classes.
Formally, the cross entropy loss of expert $E_i$ on unlabeled data $\mathbf{x}_u$ in Eq. (\ref{eq: CPE-unsup}) is re-formulated as 
{\small
\begin{align}
&L_{CPE,i}^{unsup}=\frac{1}{S}\Bigg[\ell_{CE}\bigg( E_i\Big( \text{BN}_{\text{HMT}}\big(g(\mathbf{x}^u)\big) \Big), \hat{y}_i \bigg) \nonumber\\
+&\ell_{CE}\bigg( E_i\Big( \text{BN}_{\text{MT}}\big(g(\mathbf{x}^u)\big) \Big), \hat{y}_i \bigg) \cdot \mathbb{I}\Big(\hat{y}_i\in\mathcal{M}\cup\mathcal{T}\Big)\nonumber\\
+&\ell_{CE}\bigg( E_i\Big( \text{BN}_{\text{T}}\big(g(\mathbf{x}^u)\big) \Big), \hat{y}_i \bigg) \cdot \mathbb{I}\Big(\hat{y}_i\in\mathcal{T}\Big)\Bigg].
\label{eq: unsup loss with CBN}
\end{align}
}$\mathbb{I}(\cdot)$ denotes the class mask, and $S$ equals the number of terms that used in $L_{CPE,i}^{unsup}$.
Note that we only utilize CBN in computing unsupervised loss because the class distribution of labeled set is always fixed. 
%
The illustration of CBN mechanism can be seen in Fig. \ref{fig: overview}.
We will show the effectiveness of CBN in experimental sections.

\section{Experiments}
We follow previous LTSSL works to evaluate our CPE approach on various dataset benchmarks, and make comparisons with baseline algorithms.
Also, we conduct ablation studies on CPE and provide necessary analysis.

\begin{table*}[ht]
\centering
\renewcommand\arraystretch{1.23}
\resizebox{0.78\linewidth}{!}
{
\begin{tabular}{@{}l|cccc|cc}
\toprule
Dataset & \multicolumn{4}{c|}{CIFAR-10-LT} & \multicolumn{2}{c}{CIFAR-100-LT} \\ \midrule
$N_1$ & \multicolumn{2}{c}{1500} & \multicolumn{2}{c|}{500} & \multicolumn{2}{c}{150}  \\
$M_1$ & \multicolumn{2}{c}{3000} & \multicolumn{2}{c|}{4000} & \multicolumn{2}{c}{300}\\ \midrule
$\gamma$ = $\gamma_l$ = $\gamma_u$  & 100 & 150 & 100 & 150 & 10 & 15  \\ \midrule
Supervised & 63.62\scriptsize{$\pm$0.40}  & 59.82\scriptsize{$\pm$0.32} & 47.62\scriptsize{$\pm$0.87} & 43.88\scriptsize{$\pm$1.61}  & 48.01\scriptsize{$\pm$0.45} & 45.37\scriptsize{$\pm$0.54} \\ \midrule 
FixMatch \cite{fixmatch} & 76.49\scriptsize{$\pm$0.72} & 72.15\scriptsize{$\pm$0.94} & 73.14\scriptsize{$\pm$1.03} &  65.68\scriptsize{$\pm$0.67} & 57.56\scriptsize{$\pm$0.47} & 53.97\scriptsize{$\pm$0.17}   \\
w/ DARP \cite{DARP} & 77.37\scriptsize{$\pm$0.50} & 74.02\scriptsize{$\pm$0.06} & 71.12\scriptsize{$\pm$0.82} & 65.63\scriptsize{$\pm$0.63} & 56.14\scriptsize{$\pm$0.46} & 52.81\scriptsize{$\pm$0.50}   \\
w/ CReST \cite{Crest} & 79.90\scriptsize{$\pm$0.33}  & 74.70\scriptsize{$\pm$0.53} & 77.69\scriptsize{$\pm$0.71} & 68.20\scriptsize{$\pm$0.33} & 58.56\scriptsize{$\pm$0.34}   & 55.43\scriptsize{$\pm$0.17}  \\
w/ CReST+ \cite{Crest} & 79.60\scriptsize{$\pm$0.06}  & 75.39\scriptsize{$\pm$0.42} & 78.70\scriptsize{$\pm$0.40}  & 72.73\scriptsize{$\pm$2.26} & 58.19\scriptsize{$\pm$0.37}   & 55.39\scriptsize{$\pm$0.23}   \\
w/ ABC \cite{ABC} & 84.01\scriptsize{$\pm$0.15} & 80.94\scriptsize{$\pm$0.85} & 79.40\scriptsize{$\pm$0.88}  & 69.50\scriptsize{$\pm$1.86} & 58.25\scriptsize{$\pm$0.20}  & 55.38\scriptsize{$\pm$0.47}   \\
w/ DASO \cite{DASO} & 78.87\scriptsize{$\pm$0.80} & 74.92\scriptsize{$\pm$0.36} & 73.63\scriptsize{$\pm$0.46} & 67.13\scriptsize{$\pm$1.06} & 58.16\scriptsize{$\pm$0.21}   & 54.82\scriptsize{$\pm$0.53}  \\
w/ CoSSL \cite{CoSSL} & 82.35\scriptsize{$\pm$0.79} & 79.00\scriptsize{$\pm$0.41} & 75.82\scriptsize{$\pm$0.61} & 70.56\scriptsize{$\pm$0.55} & 58.00\scriptsize{$\pm$0.39} & 55.49\scriptsize{$\pm$0.43}  \\ 
w/ SAW \cite{SAW} & 80.93\scriptsize{$\pm$0.31} & 77.67\scriptsize{$\pm$0.14} & 75.20\scriptsize{$\pm$1.01} & 68.51\scriptsize{$\pm$0.16} & 57.55\scriptsize{$\pm$0.45} & 54.00\scriptsize{$\pm$0.65}  \\ 
w/ Adsh \cite{Adsh} & 78.43\scriptsize{$\pm$0.54} & 73.96\scriptsize{$\pm$0.47} & 75.97\scriptsize{$\pm$0.68} & 66.55\scriptsize{$\pm$2.94} & 58.65\scriptsize{$\pm$0.36} & 54.55\scriptsize{$\pm$0.40}   \\ 
w/ DePL \cite{DePL} & 80.65\scriptsize{$\pm$0.52}  & 76.58\scriptsize{$\pm$0.12} & 76.98\scriptsize{$\pm$1.70} & 71.95\scriptsize{$\pm$2.54} & 57.08\scriptsize{$\pm$0.29} & 53.89\scriptsize{$\pm$0.44} \\
w/ RDA \cite{RDA} & 78.13\scriptsize{$\pm$0.27} & 72.88\scriptsize{$\pm$0.35} & 74.59\scriptsize{$\pm$0.92} & 68.59\scriptsize{$\pm$0.61} & 58.13\scriptsize{$\pm$0.45} & 54.54\scriptsize{$\pm$0.50} \\
w/ ACR \cite{ACR} & \underline{84.18\scriptsize{$\pm$0.52}} & \underline{81.81\scriptsize{$\pm$0.49}} & \textbf{81.01\scriptsize{$\pm$0.42}} & \underline{76.72\scriptsize{$\pm$1.13}} & \underline{59.83\scriptsize{$\pm$0.07}} & \underline{56.91\scriptsize{$\pm$0.09}} \\
\rowcolor{black!10}w/ \textbf{CPE (Ours)}     & \textbf{84.44\scriptsize{$\pm$0.29}} & \textbf{82.25\scriptsize{$\pm$0.34}} & \underline{80.68\scriptsize{$\pm$0.96}}  & \textbf{76.77\scriptsize{$\pm$0.53}} & \textbf{59.83\scriptsize{$\pm$0.16}} & \textbf{57.00\scriptsize{$\pm$0.51}} \\
\bottomrule
\end{tabular}%
}
\vspace{-1.5mm}
\caption{
Top-1 test set accuracy of previous LTSSL algorithms and our proposed CPE under \textit{consistent} class distributions, \textit{i.e.}, $\gamma_l=\gamma_u$, on CIFAR10-LT and CIFAR100-LT benchmarks. 
The network architecture is WRN-28-2 trained from scratch. We highlight the best number in \textbf{bold} and the second best in \underline{underline}. 
}
\label{tab:consistent-results}
\end{table*}
\begin{table*}[t!]
\centering
\renewcommand\arraystretch{1.23}
\resizebox{0.885\linewidth}{!}
{
\begin{tabular}{@{}l|cccc|c|cc}
\toprule
Dataset & \multicolumn{4}{c|}{CIFAR-10-LT} & CIFAR-100-LT & \multicolumn{2}{c}{STL-10-LT} \\ \midrule
$N_1$ & \multicolumn{2}{c}{1500} & \multicolumn{2}{c|}{500}       & 150 & \multicolumn{2}{c}{150}  \\
$M_1$ & 300         & 30         & 400  & 40 & 30  & \multicolumn{2}{c}{$\approx$100k}  \\\midrule
$\gamma_l$  & \multicolumn{4}{c|}{100}   & 10   & 10 & 20  \\ 
$\gamma_u$  & 1     & 1/100 & 1    & 1/100 & 1/10 & N/A & N/A \\ \midrule
Supervised & 63.62\scriptsize{$\pm$0.40} & 63.62\scriptsize{$\pm$0.40} & 47.62\scriptsize{$\pm$0.87} & 47.62\scriptsize{$\pm$0.87} & 48.01\scriptsize{$\pm$0.45}  & 46.85\scriptsize{$\pm$1.65} & 41.60\scriptsize{$\pm$0.77} \\ \midrule 
FixMatch \cite{fixmatch} & 73.27\scriptsize{$\pm$1.25} & 68.92\scriptsize{$\pm$0.79} & 66.47\scriptsize{$\pm$0.84} &  62.52\scriptsize{$\pm$0.93} & 57.56\scriptsize{$\pm$0.47} & 66.56\scriptsize{$\pm$1.02} & 56.29\scriptsize{$\pm$4.00} \\
w/ DARP \cite{DARP} & 73.52\scriptsize{$\pm$1.19} & 70.36\scriptsize{$\pm$1.55} & 65.80\scriptsize{$\pm$0.67} & 62.16\scriptsize{$\pm$1.10} & 56.40\scriptsize{$\pm$0.21} & 63.74\scriptsize{$\pm$0.54} & 56.03\scriptsize{$\pm$1.81} \\
w/ CReST \cite{Crest} & \underline{85.14\scriptsize{$\pm$0.19}}  & \underline{86.71\scriptsize{$\pm$0.39}} & 68.20\scriptsize{$\pm$0.33} & 76.37\scriptsize{$\pm$3.84} & 60.07\scriptsize{$\pm$0.24} & 65.52\scriptsize{$\pm$1.01} & 61.38\scriptsize{$\pm$1.75} \\
w/ CReST+ \cite{Crest} & 82.68\scriptsize{$\pm$0.93}  & 73.98\scriptsize{$\pm$1.34} & 72.73\scriptsize{$\pm$2.26} & 63.72\scriptsize{$\pm$0.87} & 59.53\scriptsize{$\pm$0.34} & 66.27\scriptsize{$\pm$0.59} & 62.63\scriptsize{$\pm$1.69} \\
w/ ABC \cite{ABC} & 81.45\scriptsize{$\pm$0.43} & 83.45\scriptsize{$\pm$0.54} & 69.50\scriptsize{$\pm$1.86} & 79.21\scriptsize{$\pm$0.44} & 59.24\scriptsize{$\pm$0.17} & 70.64\scriptsize{$\pm$0.89} & 65.68\scriptsize{$\pm$3.06} \\
w/ DASO \cite{DASO} & 76.01\scriptsize{$\pm$0.59} & 74.47\scriptsize{$\pm$0.60} & 67.54\scriptsize{$\pm$0.60} & 65.08\scriptsize{$\pm$0.90} & 59.25\scriptsize{$\pm$0.23} & 69.31\scriptsize{$\pm$0.91} & 62.45\scriptsize{$\pm$2.23} \\
w/ CoSSL \cite{CoSSL} & 79.90\scriptsize{$\pm$0.76} & 77.47\scriptsize{$\pm$0.56} & 74.31\scriptsize{$\pm$0.98} & 73.26\scriptsize{$\pm$0.78} & 57.77\scriptsize{$\pm$0.31} & 71.44\scriptsize{$\pm$0.45} & 69.01\scriptsize{$\pm$0.80} \\ 
w/ SAW \cite{SAW} & 81.50\scriptsize{$\pm$0.44} & 76.73\scriptsize{$\pm$0.66} & 68.51\scriptsize{$\pm$0.16} & 70.04\scriptsize{$\pm$1.58} & 58.12\scriptsize{$\pm$0.34} & 69.30\scriptsize{$\pm$0.69} & 65.80\scriptsize{$\pm$1.22} \\ 
w/ Adsh \cite{Adsh} & 76.55\scriptsize{$\pm$0.13} & 70.45\scriptsize{$\pm$0.70} & 66.55\scriptsize{$\pm$2.94} & 65.64\scriptsize{$\pm$1.82} & 55.49\scriptsize{$\pm$0.55} & 69.35\scriptsize{$\pm$1.12} & 64.82\scriptsize{$\pm$1.41} \\ 
w/ DePL \cite{DePL} & 72.99\scriptsize{$\pm$0.45}  & 74.53\scriptsize{$\pm$0.61} & 71.95\scriptsize{$\pm$2.54} & 69.67\scriptsize{$\pm$1.34} & 57.31\scriptsize{$\pm$0.55} & 69.46\scriptsize{$\pm$0.62} & 65.93\scriptsize{$\pm$1.22} \\
w/ RDA \cite{RDA} & 78.41\scriptsize{$\pm$0.19} & 72.62\scriptsize{$\pm$0.55} & 68.59\scriptsize{$\pm$0.61} & 68.99\scriptsize{$\pm$0.26} & 57.79\scriptsize{$\pm$0.43} & 72.63\scriptsize{$\pm$0.26} & \underline{69.24\scriptsize{$\pm$1.69}} \\
w/ ACR \cite{ACR} & 84.61\scriptsize{$\pm$0.50} & 86.29\scriptsize{$\pm$0.19} & \underline{80.10\scriptsize{$\pm$1.21}} & \underline{82.65\scriptsize{$\pm$0.31}} & \textbf{60.92\scriptsize{$\pm$0.40}} & \textbf{73.40\scriptsize{$\pm$0.73}} & 67.51\scriptsize{$\pm$1.63} \\
\rowcolor{black!10}w/ \textbf{CPE (Ours)}     & \textbf{85.86\scriptsize{$\pm$0.40}} & \textbf{87.09\scriptsize{$\pm$0.14}} & \textbf{82.32\scriptsize{$\pm$0.43}} & \textbf{83.88\scriptsize{$\pm$0.18}} & \underline{60.83\scriptsize{$\pm$0.30}} & \underline{73.07\scriptsize{$\pm$0.47}} & \textbf{69.60\scriptsize{$\pm$0.20}} \\
\bottomrule
\end{tabular}%
}
\vspace{-1.5mm}
\caption{
Top-1 test set accuracy of previous LTSSL algorithms and our proposed CPE under \textit{uniform} and \textit{inverse} class distributions, \textit{i.e.}, $\gamma_l\neq\gamma_u$, on CIFAR10-LT, CIFAR100-LT, and STL-10-LT benchmarks. 
The network architecture is WRN-28-2 trained from scratch. We highlight the best number in \textbf{bold} and the second best in \underline{underline}. 
}
\label{tab:inconsistent-results}
\end{table*}

\subsection{Experimental setting}
\noindent\textbf{Dataset}\;
We evaluate on three widely-used datasets for the LTSSL task: CIFAR-10-LT~\cite{CIFAR}, CIFAR-100-LT~\cite{CIFAR}, and STL-10~\cite{STL}.
We follow the settings in DASO~\cite{DASO} and ACR~\cite{ACR} for each dataset, detailed as below.
\begin{itemize}
    \item \textbf{CIFAR-10-LT.}\;\, We test with four settings in the \textit{consistent} case: $(N_1,\,M_1)=(1500,\,3000)$ and $(N_1,\,M_1)=(500,\,4000)$, with common imbalance ratio $\gamma$ being $100$ or $150$.
    In the \textit{uniform} case, we test with $(N_1,\,M_1)=(1500,\,300)$ and $(N_1,\,M_1)=(500,\,400)$, with $\gamma_l$ being $100$ and $\gamma_u$ being $1$.
    In the \textit{inverse} case, we test with $(N_1,\,M_1)=(1500,\,30)$ and $(N_1,\,M_1)=(500,\,40)$, with $\gamma_l$ being $100$ and $\gamma_u$ being $1/100$.

    \item \textbf{CIFAR-100-LT.}\; We evaluate on CIFAR-100-LT in both \textit{consistent} and \textit{inverse} cases.
    For the former, we set $(N_1,\,M_1)=(150,\,300)$ with $\gamma$ being $10$ or $15$.
    For the latter, $(N_1,\,M_1)=(150,\,30)$ with $(\gamma_l,\,\gamma_u)$ being $(10,\,1/10)$. 
    
    \item \textbf{STL-10-LT.}\;\, We set $N_1$ as 150 and imbalance ratio $\gamma_l$ as $10$ or $20$ for the labeled set, and directly use the original unlabeled set, which is already imbalanced and the ground-truth labels of which are not provided by the authors.
    
\end{itemize}
Note that there is no overlap between labeled and unlabeled set for all datasets.

\vspace{2mm}
\noindent\textbf{Baselines}\;
We compare with several LTSSL algorithms published in top-conferences/journals in the past few years.
These baseline algorithms include DARP~\cite{DARP}, CReST~\cite{Crest} and its variant CReST+~\cite{Crest}, ABC~\cite{ABC}, DASO~\cite{DASO}, CoSSL~\cite{CoSSL}, SAW~\cite{SAW}, Adsh~\cite{Adsh}, DePL~\cite{DePL}, RDA~\cite{RDA}, and ACR~\cite{ACR}.
For a fair comparison, we test these baselines and our CPE algorithm on the widely-used codebase USB\footnote{\url{https://github.com/microsoft/Semi-supervised-learning}}.
We use the same dataset splits and common hyper-parameters for training.

\subsection{Implementation details}
We follow the default hyper-parameters in USB and set the size of labeled and unlabeled data batch as 64 and 128, respectively.
We resize all input images to 32$\times$32.
We fix the loss weight $\lambda$ at 2, and fix the confidence threshold $\rho$ to 0.95.
We utilize the SGD optimizer with learning rate always being $3e-2$, momentum being $0.9$, and weight decay being $5e-4$.
%
%
As DNN architecture, we use WRN-28-2~\cite{wrn} without any pre-training.
We set the intensities of logit adjustment $(\tau_1,\tau_2,\tau_3)$ as $(0,2,4)$ for all datasets and experimental settings in our CPE algorithm.
For CBN mechanism in CPE, we consider the first one-third of all classes as head classes, the last one-third as tail classes, and the remaining as medium classes.
We repeat each experiment over three different random seeds, and report the mean and standard deviation of performances.
We conduct  the experiments on a single GPU of NVIDIA RTX A6000 using PyTorch \textit{v}1.10.

\subsection{Main results}
\noindent\textbf{Consistent case.}\;
We consider the \textit{consistent} case on CIFAR-10-LT and CIFAR-100-LT. 
We compare CPE and baseline algorithms based on the widely-used SSL method, FixMatch~\cite{fixmatch}.
As can be seen in Table \ref{tab:consistent-results}, our CPE method achieves better performance than all the previous baselines on CIFAR-10-LT under various settings.
For example, given $N_1=1500$, $M_1=3000$, and $\gamma=150$, CPE surpasses the previous SOTA method (ACR) by 0.44 percentage points (\textit{pp}), and all other baselines by 1.31 \textit{pp}.
On CIFAR-100-LT, the performances of CPE is on par with ACR (since the imbalance ratio reduces to 10 and the task becomes easier), but beats other baselines by $>$1.51 \textit{pp}.

\noindent\textbf{Uniform and inverse case.}\;
We consider the \textit{uniform} and \textit{inverse} case, \textit{i.e.}, $\gamma_l\neq\gamma_u$.
In Table \ref{tab:inconsistent-results}, we show that for almost all baselines, the more $\gamma_u$ differs from $\gamma_l$, the worse is the performance.
This is consistent with our findings that DNN classifier tends to predict unlabeled data as head classes, thus resulting in more erroneous when tail classes \textit{w.r.t.} labeled set are no longer tail classes in unlabeled set.
In comparison, our CPE method can consistently and largely outperforms baseline algorithms on CIFAR-10-LT, validating the effectiveness of CPE to cope with varying class distributions of unlabeled set.
Concretely, CPE surpasses ACR by 1.22 \textit{pp} and other baselines by up to 
$>$8.01 \textit{pp} with $(N_1,M_1)=(500,400)$ and $(\gamma_l,\gamma_u)=(100,1)$.
On CIFAR-100-LT and STL-10-LT, CPE achieves comparable performances with ACR, surpassing other baselines by $>$1 \textit{pp}.

\subsection{Ablation studies and analysis}

\noindent\textbf{The effect of each block.}
We examine the effectiveness of CPE in Table \ref{tab: ablation-study-by-decoupling}. We observe that the multi-expert architecture can bring significant improvements over FixMatch. 
Furthermore, the effectiveness of CBN is better reflected in the \textit{inverse} case, where our method improves over FixMatch by over 18 \textit{pp}.
Besides, in Table 4, we  observe the effects of multi-experts and CBN, showing that each block is important to reach the best results.

\begin{figure}[!t]
    \centering
    \hspace{-2mm}
    \subfloat[pseudo-label quality]{\label{fig:validation-pr}
    \includegraphics[width=0.24\textwidth]{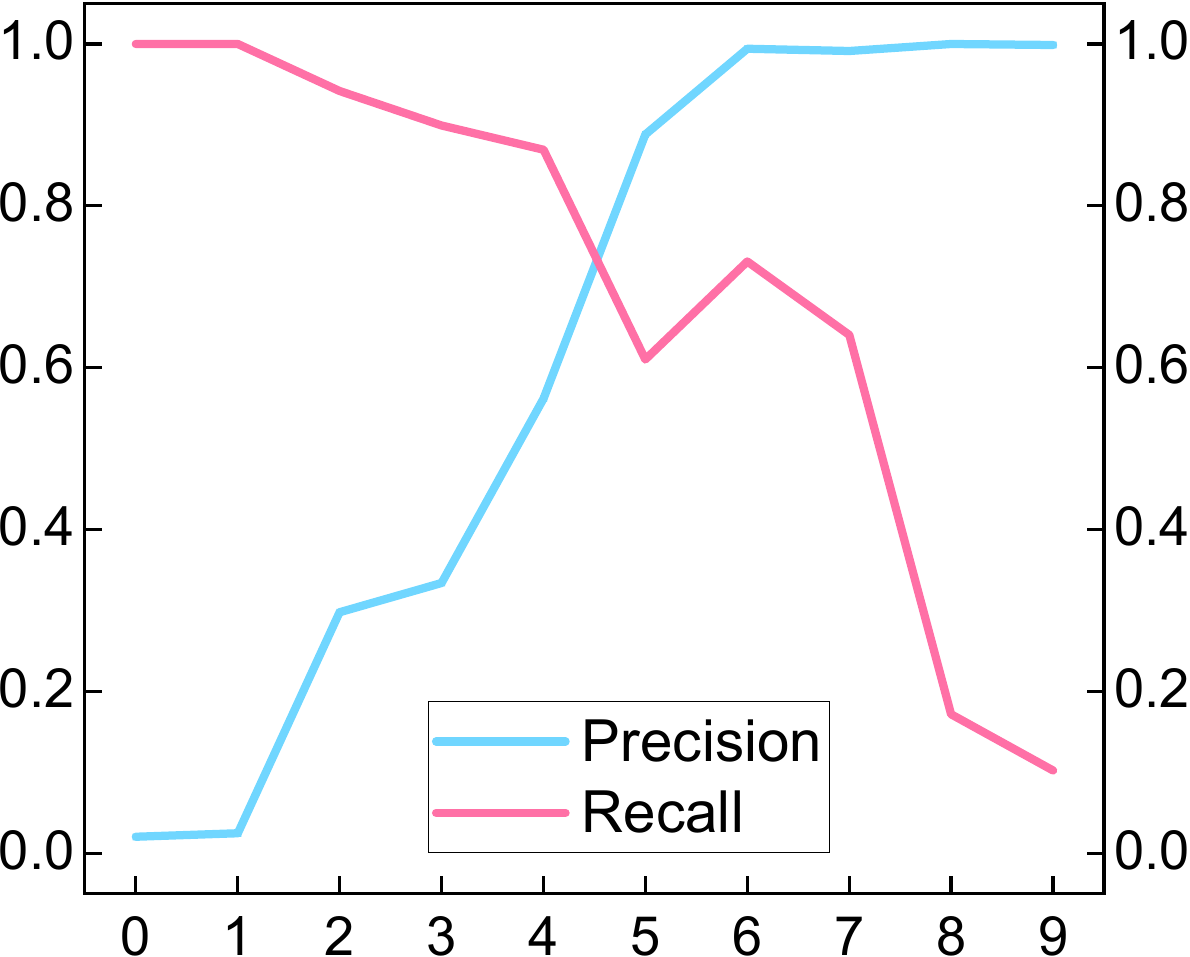}}\hspace{1mm}
    \subfloat[confusion matrix of test set]{\label{fig:confusion-matrix}
    \includegraphics[width=0.213\textwidth]{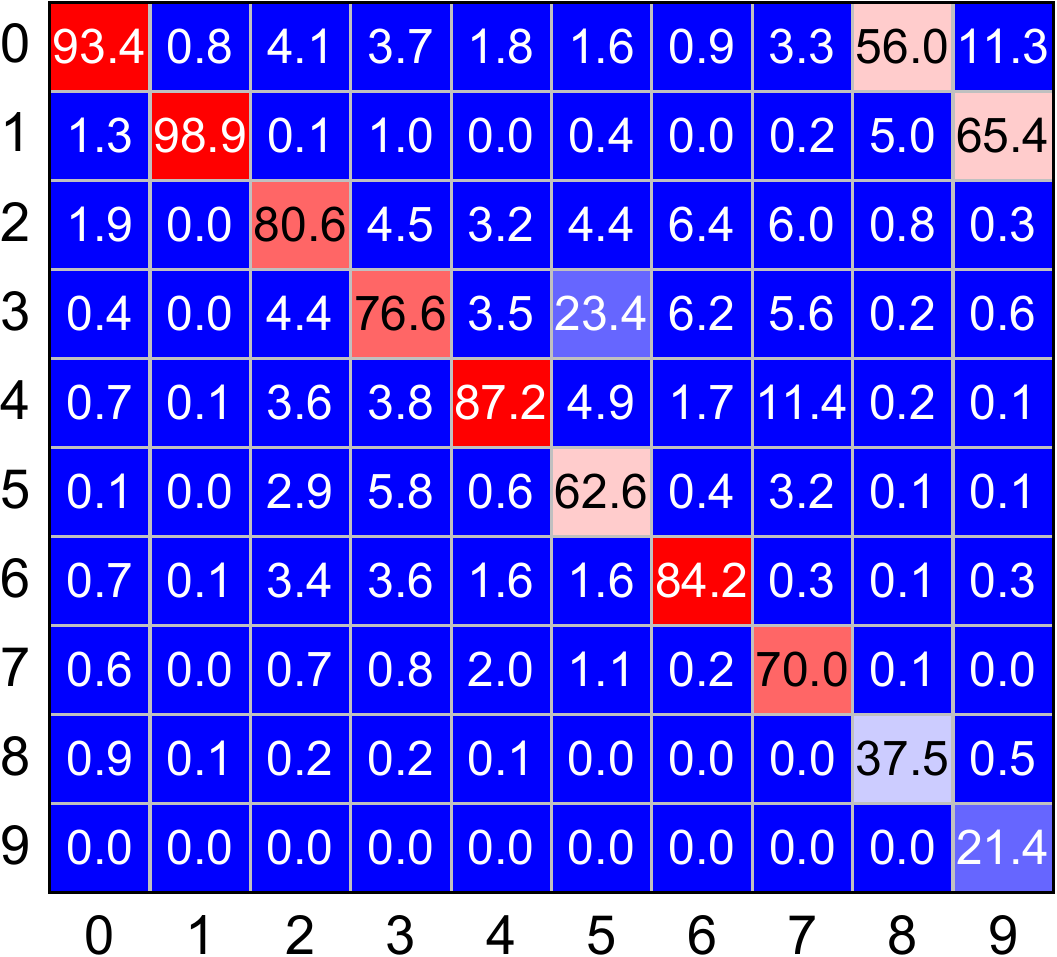}}
    \vspace{-1.5mm}
    \caption{
    \textit{Case study}: The first head trained with regular loss generates imbalanced pseudo-labels (low recall on tail classes), leads to a degeneration on the representation ability of tail classes, and finally hurt the test accuracy on tail classes.
    In contrast, low precision on head classes has much less impacts on the test accuracy. 
    }
    \vspace{-3mm}
    \label{fig: validation-experiment}
\end{figure}

\vspace{1.5mm}
\noindent\textbf{How do the low-precision pseudo-labels affect representation learning?}\;
Considering that CPE utilizes the pseudo-labels predicted by all three experts for representation learning, one may argue that CPE will incur unnecessary low-precision pseudo-labels and may harm the performance of the feature extractor. This is because only one expert can generate high-quality pseudo-labels while the others cannot (see Table \ref{tab: each-expert-does-its-work}).
Thus, we conduct an experiment to observe how the low-precision pseudo-labels affect representation learning.
Following the two-head LTSSL training paradigm, we train the first head with regular cross entropy loss, and train the second head with logit adjustment while preventing its gradient back-propagating to the shared feature extractor.
%
%
We evaluate on CIFAR-10-LT in the \textit{inverse} case with $(\gamma_l,\,\gamma_u)$ being $(100,\,1/100)$, and choose the best $\tau$ for the logit adjustment of second head via grid-search. 

As we illustrate in Fig. \ref{fig: validation-experiment}, the first head predominantly predicts unlabeled data as head classes, yielding low precision for head classes and reduced recall for tail classes.
Consequently, tail class representation capability diminishes, leading to lower tail class accuracy.
Notably, the modest precision drop for head classes minimally affects head class accuracy. 
This implies that low-precision pseudo-labels can barely harm representation learning.

\begin{figure}[t]
    \centering
    \subfloat[w/o CBN]{\label{fig:recall-E1-crop}
    {\includegraphics[width=0.22\textwidth]{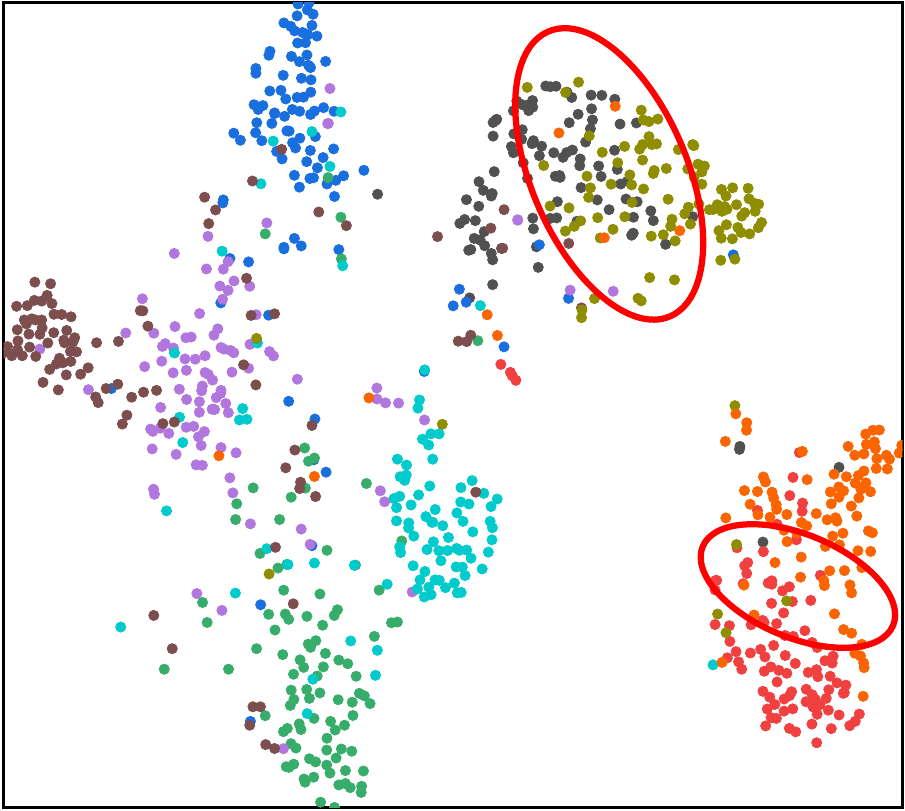}}}\hspace{1mm}
    \subfloat[w/ CBN]{\label{fig:recall-E2-crop}
    {\includegraphics[width=0.22\textwidth]{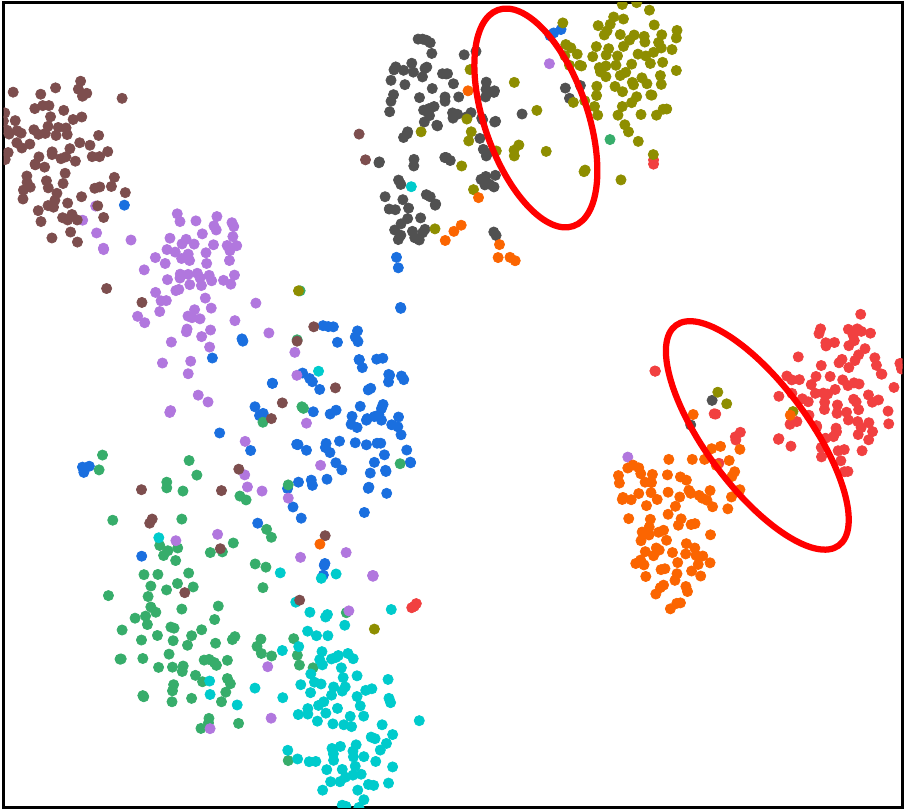}}}
    \vspace{-1.5mm}
    \caption{
    T-SNE visualization of extracted features of unlabeled data in CIFAR-10-LT with $\gamma_l=100$ and $\gamma_u=1/100$. 
    We can see that the Classwise Batch Normalization (CBN) mechanism can lead to more compact feature clusters, which can be classified easier (highlighted by red circles). 
    }
    \label{fig:decision-boundary}
\end{figure}

\vspace{1.5mm}
\noindent\textbf{Effect of CBN mechanism.}\;
As explained in Sec \ref{sec: CBN}, we propose CBN mechanism to handle varying feature distributions for CPE.
To illustrate the effect of CBN, we visualize the extracted features of all classes in Fig. \ref{fig:decision-boundary}.
As can be seen, CBN helps construct a discriminative feature space, which can boost the classification performances.

\begin{table}[!t]
\centering
\scriptsize
\renewcommand\arraystretch{1.15}
\scalebox{1.27}{
\begin{tabular}{cccc}
\hline
\multicolumn{2}{c}{FixMatch}     & \multicolumn{2}{c}{$\gamma_u$} \\
w/ Three Experts & w/ CBN  & 100      & 1/100 \\
\hline
              &            & 76.49          & 68.92            \\
\checkmark    &            & 84.02          & 86.04            \\
              & \checkmark & 79.00          & 81.22            \\
\rowcolor{black!10}\checkmark    & \checkmark & \textbf{84.79}          & \textbf{87.28}            \\ \hline
\end{tabular}
}
\vspace{-1.5mm}
\caption{Ablation on the effectiveness of multiple experts and CBN.
The dataset is CIFAR-10-LT with $(N_1,M_1)$ = $(1500,3000)$ and the imbalance ratio $\gamma_l=100$.}
\label{tab: ablation-study-by-decoupling}
\vspace{-3mm}
\end{table}

\section{Conclusion}
In this work, we have addressed the LTSSL task, focusing on the influence of pseudo-labels on representation learning. Contrary to conventional methods reliant on a single classifier for pseudo-label generation, we have addressed the challenge of adapting to varying class distributions within the unlabeled training set.
To improve pseudo-label quality, we introduce the CPE algorithm. It comprises three experts working within a shared feature extractor framework, each dedicated to modeling a distinct class distribution. We achieve this specialization by applying varying logit adjustments.
Furthermore, we introduce classwise batch-normalization that effectively handles feature distribution variance in tail classes, alleviating pseudo-labeling of additional unlabeled samples as tail classes, a common challenge in previous baselines.
We validate our proposed model's efficacy through extensive experimentation across three prominent LTSSL datasets, improving the state-of-the-art performance by up to 2.2 \textit{pp}.

\bibliography{aaai24}

\newpage
\clearpage
\appendix

In this supplementary material, we first give extra ablation studies for the logit adjustment in Section \ref{ablation}.
We then give the pseudo-code for our method in Section \ref{pseudo_code}.
Finally, in Section \ref{experimental_results} we give additional experiments in classwise comparison and time comparison with ACR \cite{ACR}, and also see if our method can work other distributions.

\section{More ablation studies} 
\label{ablation}
Due to the page limit of main manuscript, we provide more ablation studies on CPE in this section to observe the sensitivity of hyper-parameters.

\subsection{Logit adjustment intensities for multi-experts}
\begin{figure*}[!t]
    \centering
    \subfloat[$(\tau_1,\tau_2,\tau_3)=(0,2,\text{N/A})$]{\label{fig:ablation-on-three-tau-0-2}
    \includegraphics[width=0.33\textwidth]{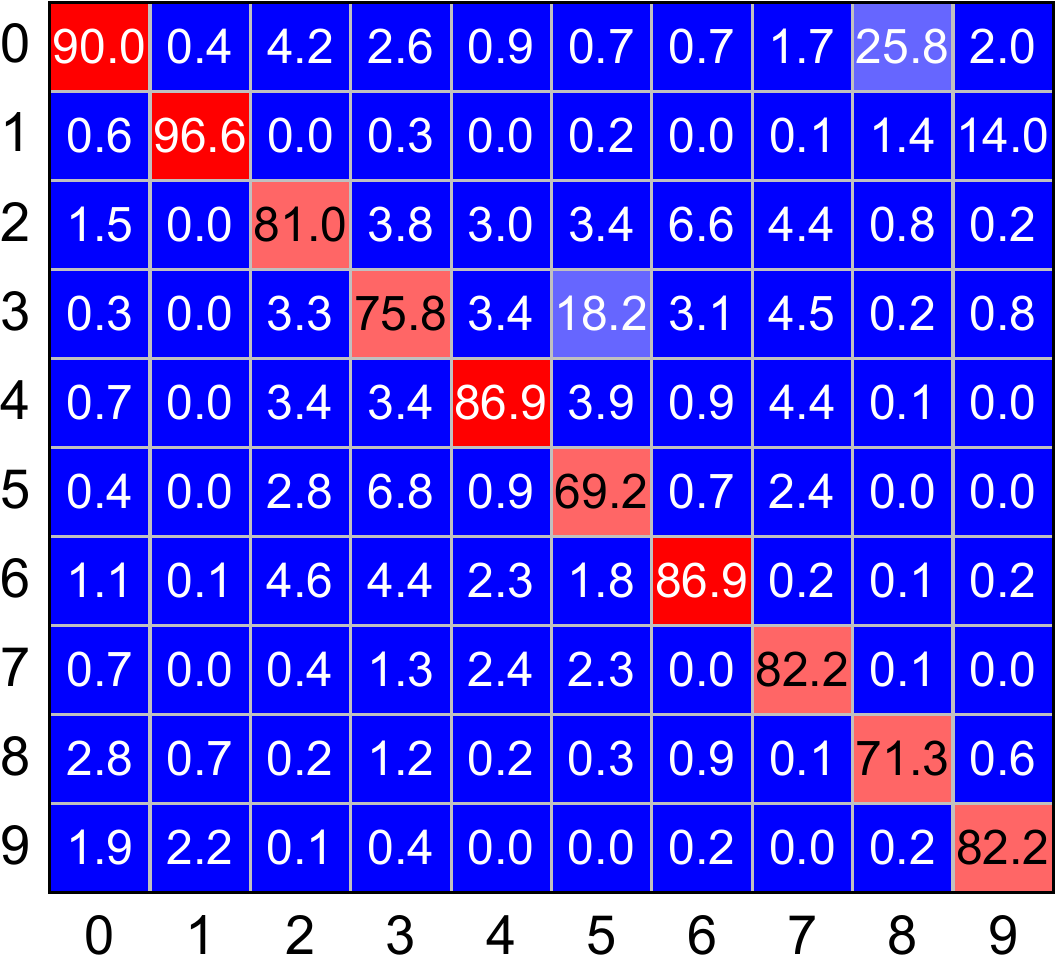}}\hspace{9mm}
    \subfloat[$(\tau_1,\tau_2,\tau_3)=(0,2,4)$]{\label{fig:ablation-on-three-tau-0-2-4}
    \includegraphics[width=0.33\textwidth]{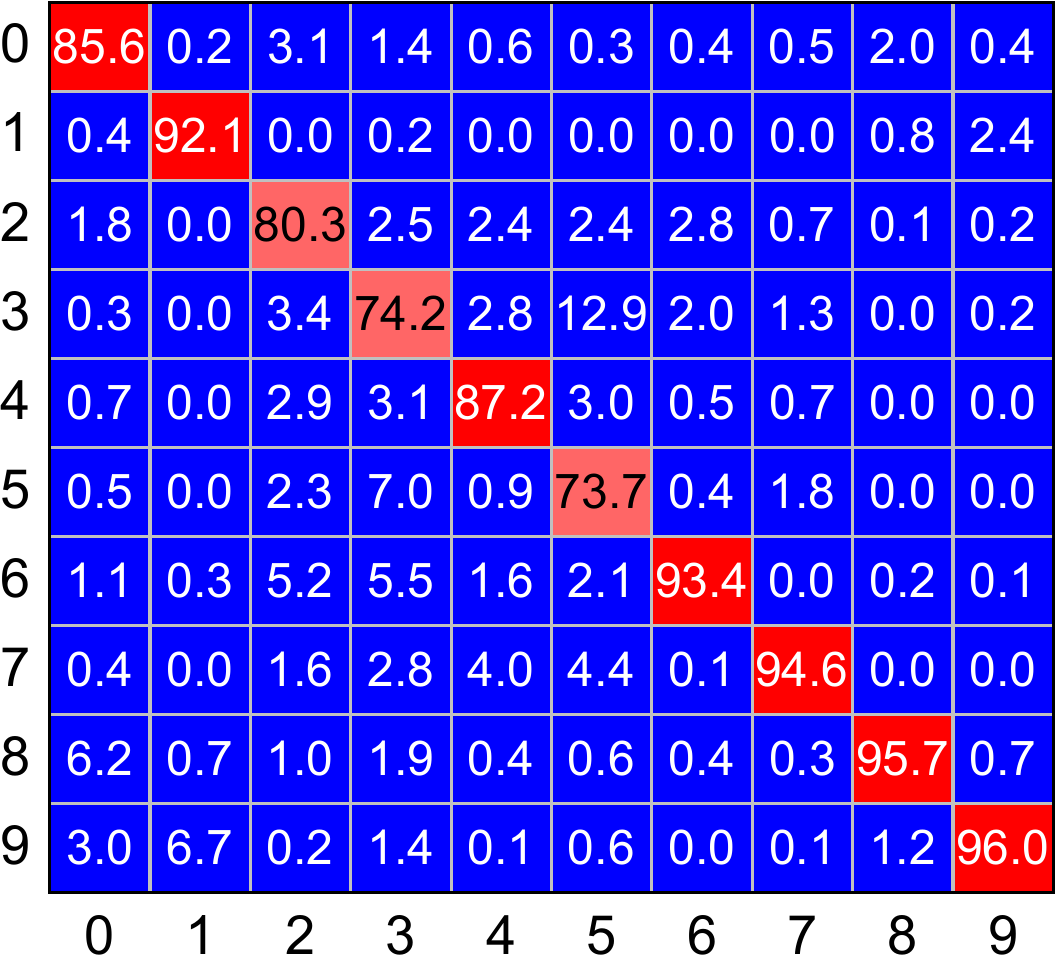}}
    \caption{
    Comparison of classwise test accuracy on different choices of $(\tau_1,\tau_2,\tau_3)$ in CPE. 
    The dataset is CIFAR-10-LT with $(N_1,M_1)$ = $(1500,3000)$ and $(\gamma_l,\gamma_u)=(100,1/100)$. 
    \textbf{(a)} two experts, with the third one being removed.
    \textbf{(b)} three proper experts (default setting in CPE).
    }
    \label{fig: ablation-on-three-tau}
\end{figure*}

The hyper-parameters $(\tau_1,\tau_2,\tau_3)$ decide the logit adjustment intensities for multi-experts to model the three different class distributions.
For example, a small set of $(\tau_1,\tau_2,\tau_3)$ can only cover the $($\textit{consistent}, \textit{less consistent}, \textit{uniform}$)$ cases, but fail to cover the \textit{inverse} case.
We test on a set of $(\tau_1,\tau_2,\tau_3)$ with the number of experts being 2 or 3 in the \textit{inverse} case, and we show the corresponding performances in Table \ref{tab: ablation-on-three-tau} and Fig. \ref{fig: ablation-on-three-tau}.
We can see that the test accuracy on tail classes significantly drops if the third expert is removed, or when $\tau_2$ and $\tau_3$ are not large enough. 
Such results emphasize the significance of multi-expert model where each expert can cover one specific class distribution and generate high-quality pseudo-labels in each corresponding case. 

\begin{table}[h]
\centering
\scriptsize
\renewcommand\arraystretch{1.1}
\scalebox{1.33}{
\begin{tabular}{cccc|c}
\hline
\multicolumn{4}{c|}{FixMatch w/ CPE} & $\gamma_u$ \\
$|E|$   & $\tau_1$ & $\tau_2$ & $\tau_3$       & 1/100      \\\hline
2 & 0.0      & 2.0      & N/A            & 82.21      \\
2 & 0.0      & 2.5      & N/A            & 83.81      \\
2 & 0.0      & 3.0      & N/A            & 84.09      \\\hline
3 & 0.0      & 1.0      & 2.0            & 85.88      \\
3 & 0.0      & 1.0      & 3.0            & 85.97      \\
3 & 0.0      & 1.5      & 3.0            & 86.88      \\
\rowcolor{black!10}3 & 0.0 & 2.0   & 4.0 & \textbf{87.28} \\\hline
\end{tabular}
}
\caption{Comparison of test accuracy on different choices of $(\tau_1,\tau_2,\tau_3)$ in CPE. 
$|E|$ denotes the number of experts. 
The dataset is CIFAR-10-LT with $(N_1,M_1)$ = $(1500,3000)$ and the imbalance ratio $\gamma_l=100$.
The default setting in our experiments is three experts with $(0.0,\,2.0,\,4.0)$. 
}
\label{tab: ablation-on-three-tau}
\end{table}

\subsection{Use all three experts or just $E_2$ for inference?}
The three heads are experts in skewed/balanced/inversely-skewed distributions.
Since the test set distribution is usually balanced, CPE only utilizes $E_2$ during testing.
In fact, we tried to utilize the average of three predictions for testing (see Table \ref{tab: only E2}), and the performance slightly drops.
The reason is that $E_1$ and $E_3$ yield too large probabilities on head and tail classes, respectively, and hurt the performance on medium classes.

\begin{table}[!t]
\centering
\renewcommand\arraystretch{1.3}
\scalebox{0.92}
{
\begin{tabular}{c c c}
\toprule
Classifier          & 100   & 1/100 \\ \midrule
$E_1$               & 78.88 & 84.24 \\
\rowcolor{black!10}$E_2$ (CPE)               & \textbf{84.79} & \textbf{87.28} \\
$E_3$               & 59.23 & 81.38 \\
Averaging three heads  & 84.45 & 87.16 \\
\bottomrule
\end{tabular}%
}
\caption{
Comparison of test accuracies among three experts.
The dataset is CIFAR-10-LT with $(N_1,M_1)$ = $(1500,3000)$, $\gamma_l$ being 100, and $\gamma_u$ being 100 or 1/100.
}
\label{tab: only E2}
\end{table}

\begin{table}[!t]
\centering
\scriptsize
\renewcommand\arraystretch{1.1}
\scalebox{1.33}{
\begin{tabular}{l|c}
\hline
Algorithm            & Per-epoch time (s) \\\hline
ACR                  & 84       \\
CPE \textbf{w/o} CBN & 86 (+2.4\%)      \\ 
\rowcolor{black!10}CPE \textbf{w/} CBN  & 86 (+2.4\%)      \\\hline
\end{tabular}
}
\caption{Comparison of per-epoch training time between ACR and CPE. 
Figures in bracket denote the relative increment compared with ACR.}
\label{tab: time-cost-comparison}
\end{table}

\section{Pseudo-code of our CPE algorithm}
\label{pseudo_code}
In Algorithm \ref{alg:algorithm}, we provide the pseudo-code of CPE which consists of multi-experts and the CBN mechanism.

\begin{algorithm}[!t]
\caption{Pseudo-code of CPE algorithm}
\label{alg:algorithm}
\textbf{Input}: Labeled training set $\mathcal{D}_l$, unlabeled training set $\mathcal{D}_u$, feature extractor $g$, three experts $\{E_1,E_2,E_3\}$, class distribution of labeled set $\pi$, intensities of logit adjustment $\{\tau_1,\tau_2,\tau_3\}$, number of training iterations $T$, strong data augmentation $\mathcal{A}$.\\
\textbf{Output}: $g$ and $\{E_1,E_2,E_3\}$.
\begin{algorithmic}[1] 
\STATE \textbf{for} $t = 1,\dots,T$ \textbf{do}
\STATE \quad Sample a labeled data batch $\{(\mathbf{x}^l,y)\}$ from $\mathcal{D}_l$ and an unlabeled data batch $\{\mathbf{x}^u\}$ from $\mathcal{D}_u$.
\STATE \quad Compute supervised loss with $g$ and $E_1$ on $\{(\mathbf{x}^l,y)\}$.
\STATE \quad Compute supervised loss with $g$ and $E_2$ on $\{(\mathbf{x}^l,y)\}$.
\STATE \quad Compute supervised loss with $g$ and $E_3$ on $\{(\mathbf{x}^l,y)\}$.
\STATE \quad Generate pseudo-label $\hat{y}_1$ for $\{\mathbf{x}^u\}$ with $g$ and $E_1$.
\STATE \quad Generate pseudo-label $\hat{y}_2$ for $\{\mathbf{x}^u\}$ with $g$ and $E_2$.
\STATE \quad Generate pseudo-label $\hat{y}_3$ for $\{\mathbf{x}^u\}$ with $g$ and $E_3$.
\STATE \quad Compute unsupervised loss on $\{(\mathbf{x},\hat{y}_1)\}$ for $g$ and $E_1$.
\STATE \quad Compute unsupervised loss on $\{(\mathbf{x},\hat{y}_2)\}$ for $g$ and $E_2$.
\STATE \quad Compute unsupervised loss on $\{(\mathbf{x},\hat{y}_3)\}$ for $g$ and $E_3$.
\STATE \quad Update $g$ and $\{E_1,E_2,E_3\}$.
\STATE \textbf{end for}
\end{algorithmic}
\end{algorithm}

\begin{figure*}[!t]
    \centering
    \subfloat[F1 score in \textit{uniform}]{\label{fig:classwise-F1-uniform}
    \includegraphics[width=0.23\textwidth]{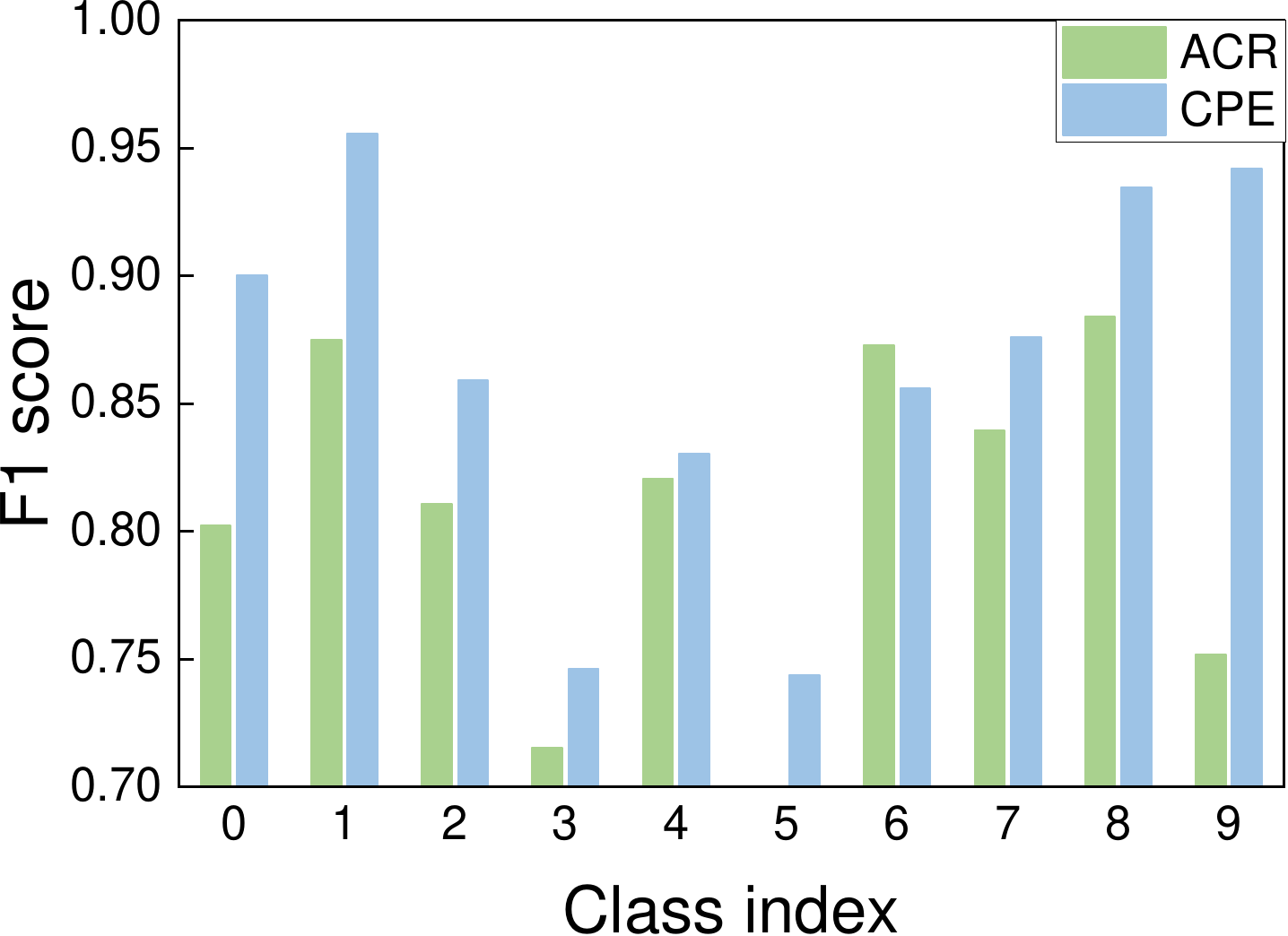}}\hspace{1mm}
    \subfloat[Test accuracy in \textit{uniform}]{\label{fig:classwise-ACC-uniform}
    \includegraphics[width=0.23\textwidth]{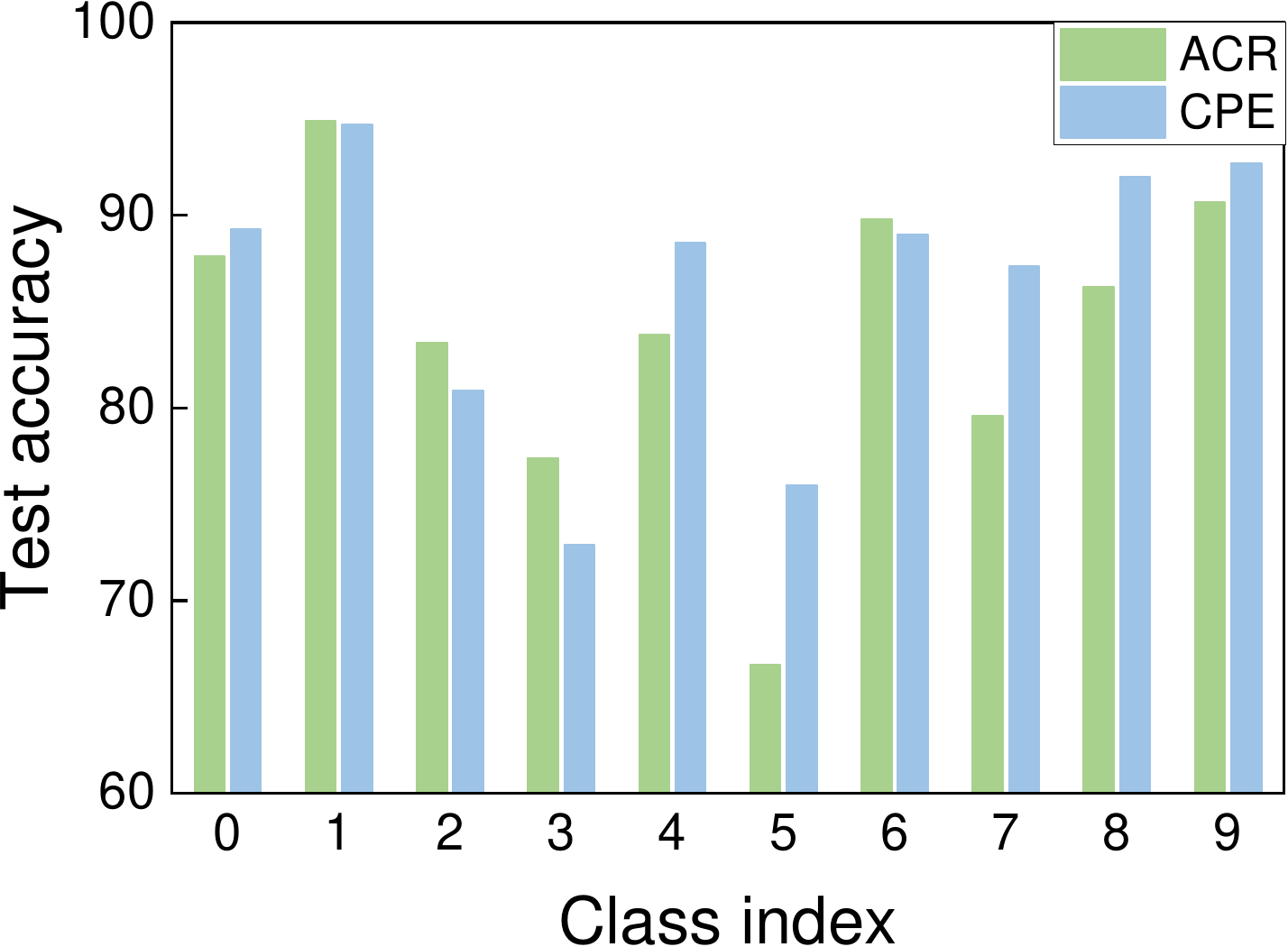}}\hspace{1mm}
    \subfloat[F1 score in \textit{inverse}]{\label{fig:classwise-F1-inverse}
    \includegraphics[width=0.23\textwidth]{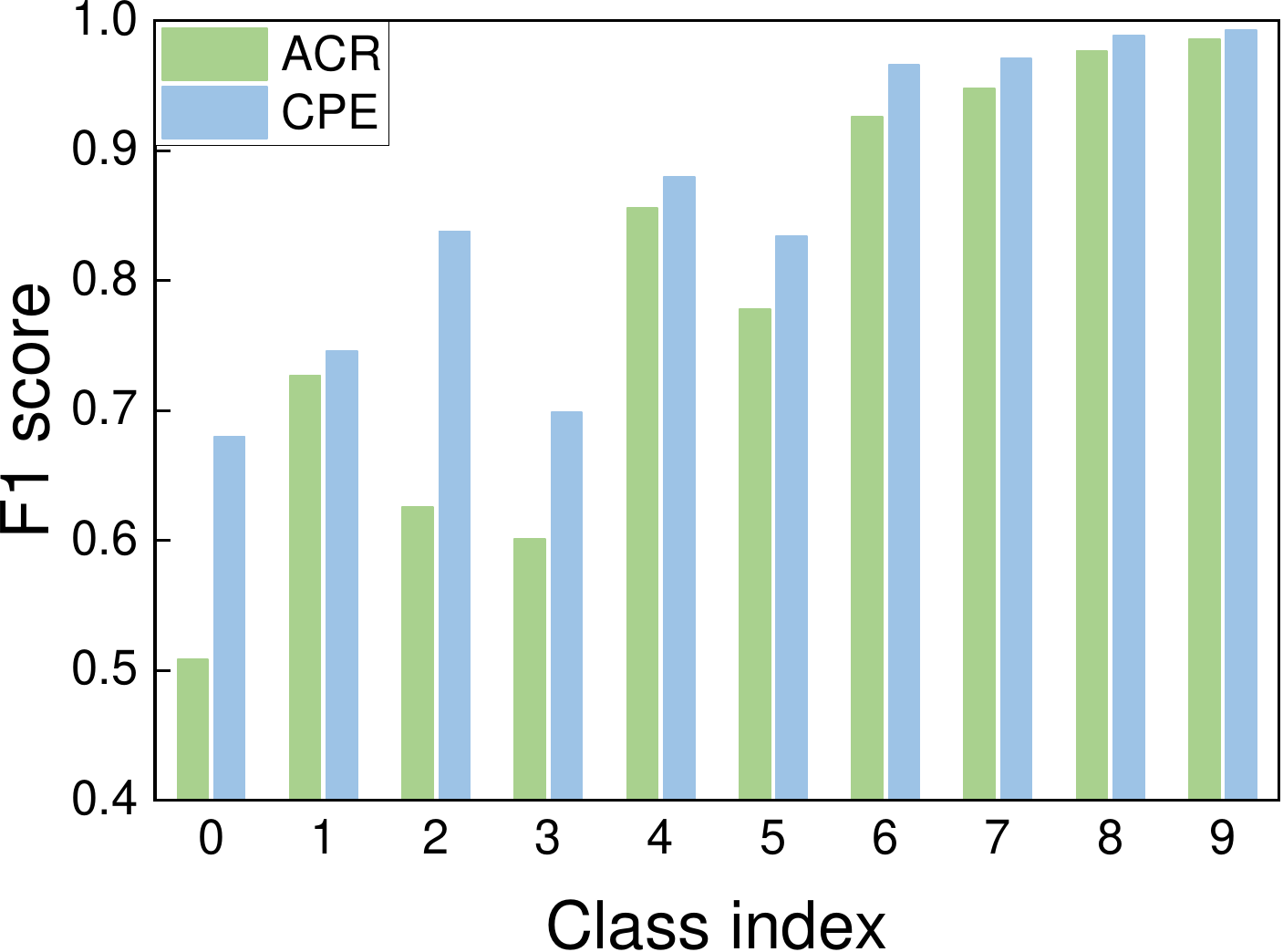}}\hspace{1mm}
    \subfloat[Test accuracy in \textit{inverse}]{\label{fig:classwise-ACC-inverse}
    \includegraphics[width=0.23\textwidth]{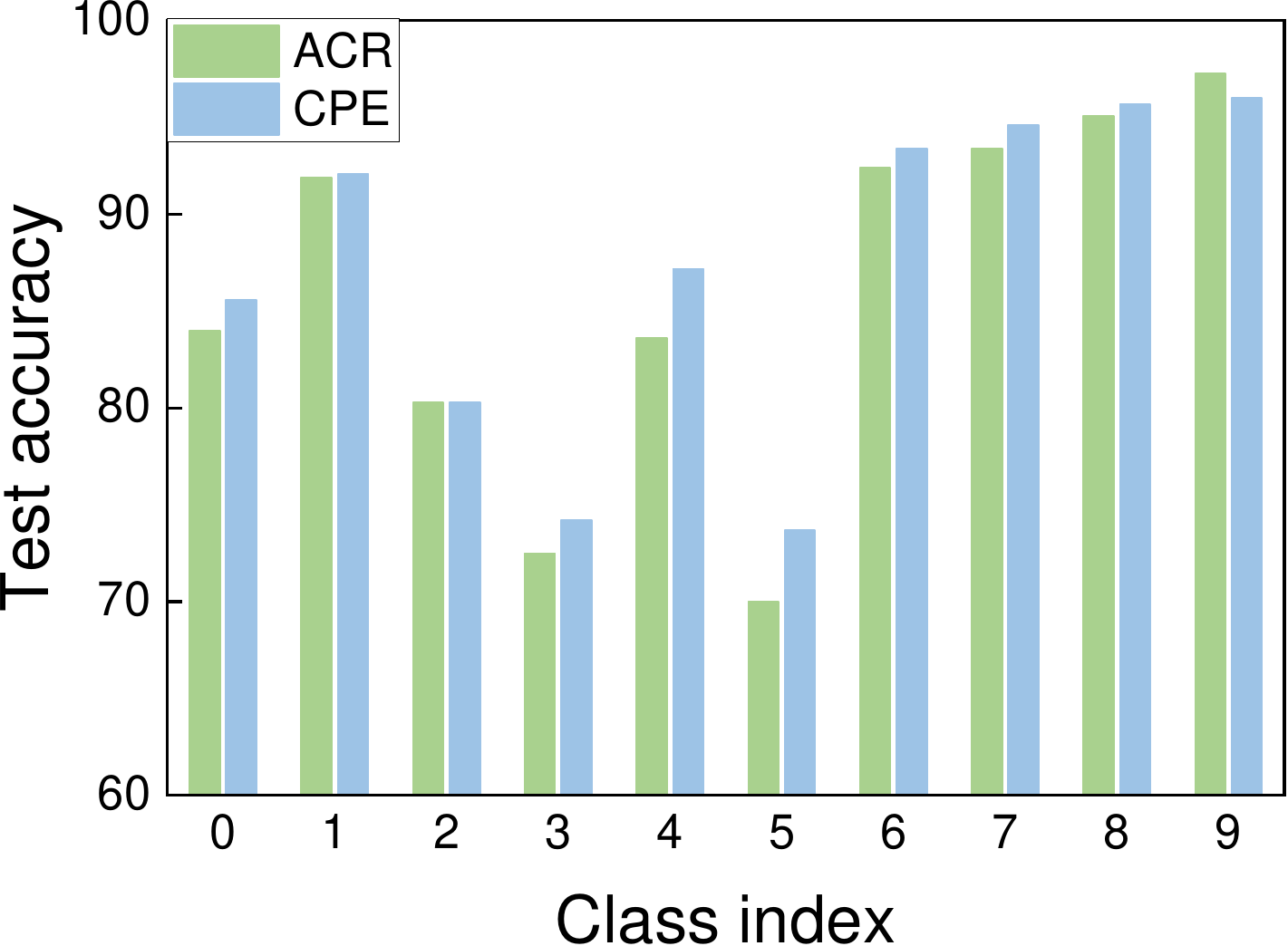}}
    \caption{
    Classwise comparisons on F1 score of pseudo-labels and test accuracy between ACR and our CPE. The dataset is CIFAR-10-LT with $(N_1,M_1)$ = $(1500,3000)$, the imbalance ratio $\gamma_l=100$, and $\gamma_u=1$ (\textit{uniform}) or $\gamma_u=1/100$ (\textit{inverse}).
    }
    \label{fig: classwise-comparison}
\end{figure*}

\section{More experimental results}
\label{experimental_results}
We now display more experimental results and make comparison with ACR. 


\subsection{Classwise comparison with ACR}
We record the F1 score of pseudo-labels within each class during training, and the final test accuracy on each class.
We evaluate on \textit{uniform} and \textit{inverse} cases and make comparison with ACR. 
We can observe from Fig. \ref{fig: classwise-comparison} that:
\textbf{(a)} In each class, the test accuracy shows a similar trend with F1 score of pseudo-label. A higher F1 score usually indicates a higher test accuracy.
\textbf{(b)} Our CPE can generate pseudo-labels with higher quality and surpass the performance of ACR at most of time.

\subsection{Time cost comparison with ACR}
Although our CPE algorithm trains more than one expert, in Table \ref{tab: time-cost-comparison}, we show that CPE does not increase the training time significantly. In fact, compared to ACR, our method comes with only a 2.4\% time increment.
Additionally, integrating the CBN mechanism to CPE does not bring additional training time cost (only a 2.4\% time increment compared to ACR).

\subsection{Does CPE work with other distributions?}
We compare CPE with previous SOTA method ACR on four new distributions (see Table \ref{tab: more distributions}), where ``shuffle \#1" and ``shuffle \#2" mean that we randomly shuffle the number of samples within each class when $\gamma_u=1/100$, \textit{e.g.}, $M_C>M_1>M_{C-1}>M_2>\cdots$. We see that CPE still beats ACR.

\begin{table}[!t]
\centering
\renewcommand\arraystretch{1.3}
\scalebox{0.92}
{
\begin{tabular}{c | c c c c}
\toprule
$\gamma_u$ & 1/50 & 1/150 & shuffle \#1 & shuffle \#2   \\ \midrule
ACR & 87.57 & 85.76 & 84.33 & 85.47 \\ 
\rowcolor{black!10}CPE & \textbf{88.04} & \textbf{86.57} & \textbf{84.75} & \textbf{85.96} \\
\bottomrule
\end{tabular}%
}
\caption{
Comparison with ACR under various distributions in CIFAR-10-LT with $(N_1,\gamma_l)$ = $(1500,100)$.
}
\vspace{-1mm}
\label{tab: more distributions}
\end{table}

\section{Connection to recent methods}
Also based on multiple experts, SADE~\cite{SADE} learns towards three simulated class distributions for test-agnostic long-tailed setting. At test time, SADE aggregates the softmax predictions of all experts in a self-supervised manner for a better accuracy. 
Afterwards, BalPoE~\cite{BalPoE} improves SADE and is proved to achieve calibrated and unbiased predictions.
With respect to each training sample, we note that the predicted softmax scores of all experts are always learned towards a same class label. In contrast, our CPE method allows each expert to learn towards each guessing on unlabeled sample, thus one of the three experts can always yield high-quality pseudo-labels.

\end{document}